\def\year{2021}\relax
\documentclass[letterpaper]{article} %

\usepackage{aaai21_no_copyright}  %
\usepackage{soul}
\usepackage{times}  %
\usepackage{helvet} %
\usepackage{courier}  %
\usepackage[hyphens]{url}  %
\usepackage{graphicx} %
\usepackage{caption} %
\graphicspath{{../}{./}{./figures/}}

\usepackage[utf8]{inputenc}
\usepackage{graphicx}
\usepackage{xcolor}
\usepackage[justification=centering]{subfig}

\usepackage{amsmath}
\usepackage{amsfonts}
\usepackage{amssymb}
\usepackage{mathtools}
\usepackage{todonotes}
\usepackage{enumitem}
\usepackage{comment}

\usepackage{tikzpagenodes}

\usepackage{algorithm,algorithmic}

\newcommand{\phdone}[1]{} %
\newcommand{\ncomment}[1]{}
\newcommand{\qcomm}[1]{{\color{red}Quest.}} %

\newcommand{\argmax}{\mathop{\mathrm{argmax}}}

\setlist[enumerate]{topsep=0pt,itemsep=-1ex,partopsep=1ex,parsep=1ex}

\title{
    PLGRIM: Hierarchical Value Learning for \\Large-scale Autonomous Exploration in Unknown Environments
}
\author{
		Sung-Kyun Kim,\thanks{These authors contributed equally to this work. \newline \hspace*{1.1em} \copyright2021. All rights reserved.}\textsuperscript{\rm 1}  %
    Amanda Bouman,$^{*}$\textsuperscript{\rm 2}
    Gautam Salhotra,\textsuperscript{\rm 3}
    David D. Fan,\textsuperscript{\rm 1} \\
    Kyohei Otsu,\textsuperscript{\rm 1}
    Joel Burdick,\textsuperscript{\rm 2}
    Ali-akbar Agha-mohammadi\textsuperscript{\rm 1} \\
}
\affiliations{
    \textsuperscript{\rm 1} NASA Jet Propulsion Laboratory, California Institute of Technology, \\
    \textsuperscript{\rm 2} Department of Mechanical and Civil Engineering, California Institute of Technology, \\
    \textsuperscript{\rm 3} Department of Computer Science, University of Southern California \\
    \{sung.kim, david.d.fan, kyohei.otsu, aliagha\}\!@jpl.nasa.gov, \\
    abouman@caltech.edu,
    jwb@robotics.caltech.edu,
    salhotra@usc.edu
}

\begin{document}

\maketitle

\begin{abstract}
In order for an autonomous robot to efficiently explore an unknown environment, it must account for uncertainty in sensor measurements, hazard assessment, localization, and motion execution.
Making decisions for maximal reward in a stochastic setting requires value learning and policy construction over a belief space, i.e., probability distribution over all possible robot-world states.
However, belief space planning in a large spatial environment over long temporal horizons suffers from severe computational challenges.
Moreover, constructed policies must safely adapt to unexpected changes in the belief at runtime.
This work proposes a scalable value learning framework, PLGRIM (Probabilistic Local and Global Reasoning on Information roadMaps), that bridges the gap between \textit{(i)} local, risk-aware resiliency and \textit{(ii)} global, reward-seeking mission objectives.  
Leveraging hierarchical belief space planners with information-rich graph structures, PLGRIM addresses large-scale exploration problems 
while providing locally near-optimal coverage plans. 
We validate our proposed framework with high-fidelity dynamic simulations in diverse environments and on physical robots
in Martian-analog lava tubes.
\end{abstract}

\section{Introduction}\label{sec:intro}

\phdone{High-level mission}
Consider a large-scale coverage mission in an unknown environment, in which a robot is tasked with exploring and searching a GPS-denied unknown area, under given time constraints. This problem has a wide range of applications, such as inter-planetary exploration and search-and-rescue operations \cite{blank2020robotic,nagatani2013emergency}. %
Essential elements of an autonomy architecture needed to realize such a mission include creating a map of the environment, accurately predicting risks, and planning motions that can meet the coverage and time requirements while minimizing risks.  In such an architecture, quantifying and planning over uncertainty is essential for creating robust, intelligent, and optimal behaviors.

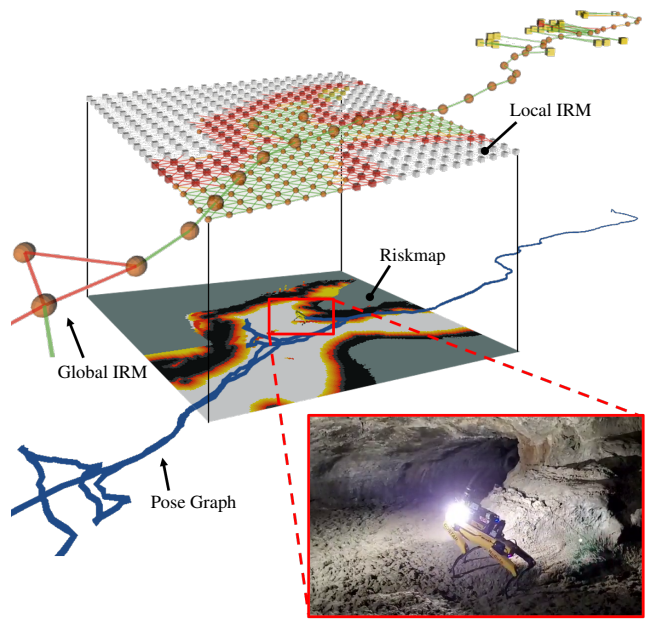
\begin{figure}[t!]
\centering
    \begin{tikzpicture}
    \node[anchor=south west,inner sep=0] (image) at (0,0) {\includegraphics[width=1\columnwidth]{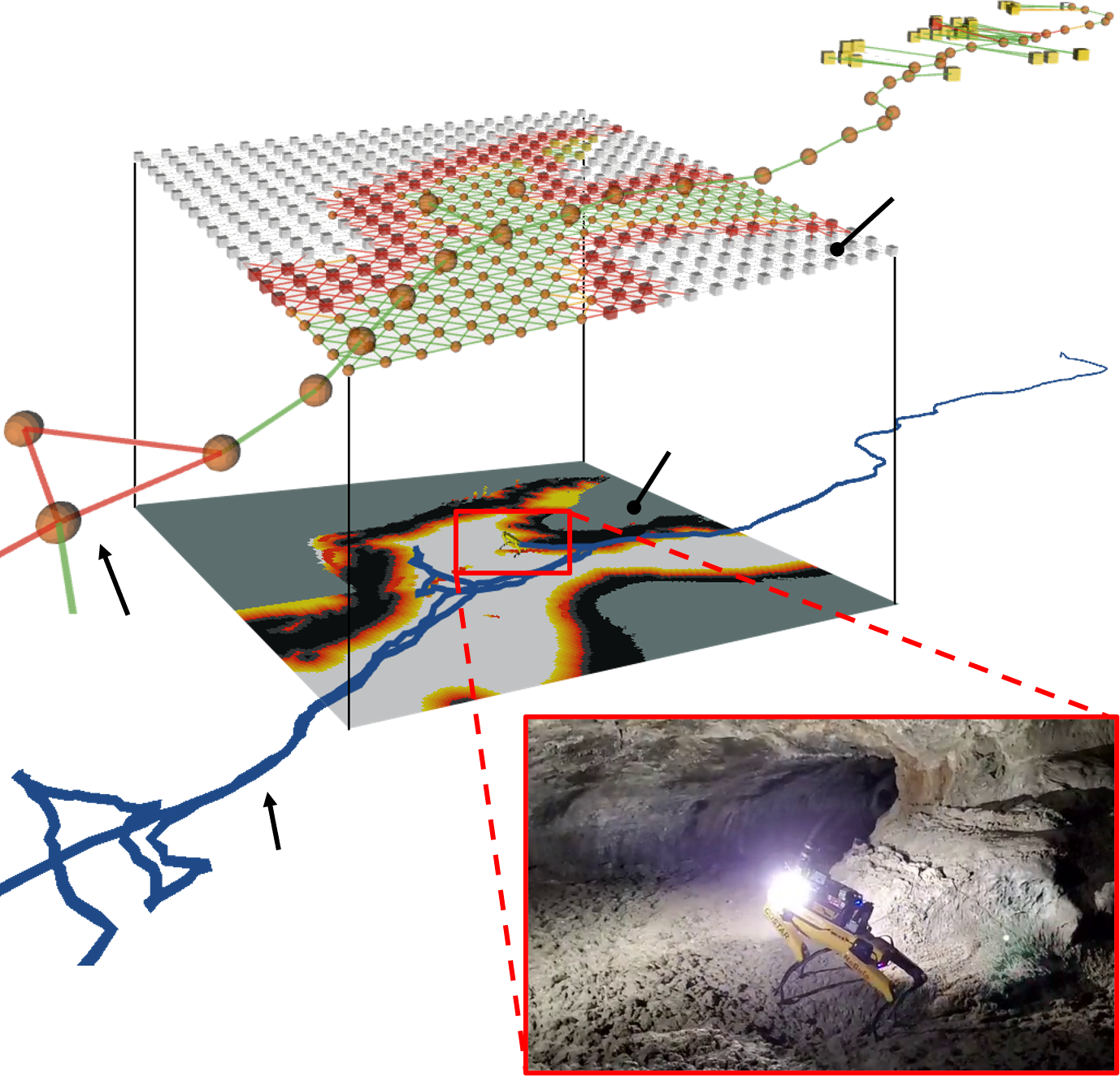}};
	    \begin{scope}[x={(image.south east)},y={(image.north west)}]

	    	\node [font=\scriptsize,above left,align=right,black] at (0.93,0.81) {Local IRM}; %
	    	\node [font=\scriptsize,above left,align=right,black] at (0.7,0.565) {Riskmap}; %
	    	\node [font=\scriptsize,above left,align=right,black] at (0.37,0.16) {Pose Graph};
	    	\node [font=\scriptsize,above left,align=right,black] at (0.23,0.38) {Global IRM};

	    \end{scope}
	\end{tikzpicture}	
  \caption{Hierarchical Information RoadMaps (IRMs) generated during Au-Spot's autonomous exploration of Martian-analog caves at Lava Beds National Monument, Tulelake, CA.} \label{fig:IRMs} 
\end{figure}

From a value learning perspective, a coverage planning problem in an unknown space can be considered an active learning problem over the robot's belief, where belief is defined as the probability distributions over all possible joint robot-world states.
The objective is to find the best action sequence that maximizes the accumulated reward over time.  The agent must accumulate data to incrementally build a model of its environment, and need to understand the effects of its actions on the quality and quantity of data it collects.

\phdone{Problem description--POMDP perspective}
Since the agent's future actions affect its belief of the world and robot state, this coverage problem is fundamentally a Partially Observable Markov Decision Process (POMDP) problem~\cite{pomdps_monahan1982}.
The agent employs the underlying intrinsic model of the sequential action-observation process under uncertainty, so that it %
expands its search structure over the belief space and learns the value in a more sample-efficient manner than model-free learning approaches.
In addition, its non-myopic reasoning can provide better performance than frontier-based exploration approaches with one-step look-ahead.

\phdone{Gap in the state-of-the-art}
Belief value learning in the POMDP setting intrinsically suffers from the curse of dimensionality \cite{KLC98} and curse of history \cite{Pineau03}. Many powerful methods are proposed to extend the spatial and temporal horizons of POMDPs with varying degrees of efficiency and accuracy, such as \cite{silver2010monte,somani2013despot,bonet1998learning,kim2019pomhdp}. In this paper, we focus on challenging exploration problems with very large spatial extents ($>\!\!1$~km), long temporal horizons ($>\!\!1$~hour), and high dimensional belief states (including beliefs on the state of the environment) that exacerbate the curses of dimensionality and history for POMDPs. %

\phdone{Contributions}
The main contribution of this work %
is in three-fold:
\begin{enumerate}[label={\arabic*)}]
\item Scalable belief representation of local traversability and global coverage states of large environments. %
	\item Hierarchical value learning for efficient coverage policy search over a long horizon under uncertainty.
	\item Policy reconciliation between planning episodes for adaptive and resilient execution in the real world.
\end{enumerate}
More precisely, we introduce spatial and temporal approximations of the coverage policy space to enable computational tractability for real-time online solvers.
Spatially, we decompose the belief space into task-relevant partitions of the space,
enriched with environment map estimates. %
The partitioning structure is called an Information Roadmap (IRM) as shown in Fig.~\ref{fig:IRMs} \cite{Ali14-IJRR}.
Temporally, we decompose the problem into local and global hierarchical levels and solve for belief space policies that provide locally near-optimal coverage plans with global completeness.
We then propose a Receding Horizon Planning (RHP)-based technique to address real-world stochasticity in state estimation and control at runtime.

\phdone{Outline}
The remainder of this paper is as follows: following the related work discussion, Section~\ref{sec:formulation} formalizes the unknown environment coverage problem. In Section~\ref{sec:plgrim}, we propose a hierarchical belief representation and value learning framework. Experimental results in simulation and on a physical robot are presented in Section~\ref{sec:exp_results}, and Section~\ref{sec:conclusion} concludes this paper.

\section{Related Work}\label{sec:related_work}
\phdone{Coverage--Frontier-based exploration}
Frontier-based exploration is a widely used approach for autonomous exploration (e.g., \cite{yamauchi1997frontier,tao2007motion,keidar2012robot,heng2015efficient,gonzalez2002navigation,grabowski2003autonomous}). By continuing exploration until exhausting all remaining frontiers, frontier-based approaches can guarantee completeness of the coverage of reachable spaces.  These methods typically rely on myopic (e.g., one-step) look-ahead greedy policies, selecting the best frontier upfront. Hence, they can be subject to local minima and provide suboptimal solutions in time.

\phdone{Coverage--(Model-free) RL-based approaches}
Model-free reinforcement learning (RL) has been applied to coverage and exploration problems (e.g., \cite{pathak_icm,burda2018study,rnd,ECR2018}). In this setting, the typical approach is to find a policy which maps sensor data to actions, with the objective of maximizing the reward. When it comes to long-range, large-scale, and safety-critical missions on physical robots, collecting necessary data can be a significant challenge for this class of methods.

\phdone{Coverage--(Model-based RL) POMDP approaches}
POMDP-based approaches generate a non-myopic policy by considering long-horizon action sequences (e.g., \cite{kurniawati2011motion}, \cite{bai2015intention}), interactively learning the value function, and returning the best action sequence that maximizes the accumulated rewards. Different methods have reduced the complexity of the POMDP problem in coverage and exploration problems. \cite{indelman2015planning} and \cite{martinez2009bayesian} employed a direct policy search scheme with a Gaussian belief assumption. \cite{Lauri2016planning} extended this to non-Gaussian beliefs using the POMCP (Partially Observable Monte-Carlo Planning) solver. %
However, when it comes to the large-scale coverage missions, %
the current approaches do not scale well due to the curse of history and dimensionality \cite{Pineau03}.

\phdone{Large scale--Hierarchical approaches}
Hierarchical planning structures \cite{kaelbling2011planning} aim to tackle larger problems by employing multiple solvers running at different resolutions, and are found to be effective.  
In the coverage and exploration context, \cite{umari2017autonomous} applied hierarchical planning to frontier-based exploration, while  \cite{dang2019explore} extended the lower-level module to a more sophisticated frontier selection algorithm which considers the information gain along each path. \cite{Lauri2016planning} replaced the lower-level module with a POMDP-based planner to improve local coverage performance with non-myopic planning. \cite{kim2019bi} proposed a hierarchical online-offline solver for risk-aware navigation. \cite{vien2015hierarchical} suggested a hierarchical POMCP framework which outperformed Bayesian model-based hierarchical RL approaches in some benchmarks.

\section{Problem Formulation}
\label{sec:formulation}

Autonomous exploration in unknown environments under motion and sensing uncertainty can be formulated as a Partially Observable Markov Decision Process (POMDP), which is one of the most general models for sequential decision making.
In this section, we present a POMDP formulation for coverage problems and address its intrinsic challenges.

\subsection{Preliminaries}
\phdone{POMDP Elements}
A POMDP is described as a tuple $\langle \mathbb{S}, \mathbb{A}, \mathbb{Z}, T, O, R \rangle$, where $\mathbb{S}$ is the set of joint robot-and-world states, $\mathbb{A}$ and $\mathbb{Z}$ are the set of robot actions and observations.
At every time step, the agent performs an action $a \in \mathbb{A}$ and receives an observation $z \in \mathbb{Z}$ resulting from the robot's perceptual interaction with the environment.
The motion model $T(s, a, s') = p(s'\,|\,s, a)$ defines the probability of being at state $s'$ after taking action $a$ at state $s$.
The observation model $O(s, a, z) = p(z\,|\,s, a)$ is the probability of receiving observation $z$ after taking action $a$ at state $s$.
The reward function $R(s, a)$ returns the expected utility for executing action $a$ at state $s$.
Belief state $b_t \in \mathbb{B}$ at time $t$ denotes a posterior distribution over states conditioned on the initial belief $b_0$ and past action-observation sequence, i.e., $b_{t} = p(s \,|\, b_0, a_{0:t-1}, z_{1:t})$.

\phdone{POMDP Objective function}
The optimal policy of a POMDP for all time $t \in [0,\infty)$, $\pi_{0:\infty}^* \! : \mathbb{B} \to \mathbb{A}$, is defined as:
\begin{align}
  \pi_{0:\infty}^*(b) &= \argmax_{\pi \in \Pi_{0:\infty}} \, \mathbb{E} \sum_{t=0}^{\infty} \gamma^t r(b_t, \pi_t(b_t)),
  \label{eq:objective_function}
\end{align}
where $\gamma \in (0, 1]$ is a discount factor for the future rewards, $\Pi_{0:\infty}$ is the space of possible policies, and $r(b,a)=\int_s R(s,a)b(s)\mathrm{d}s$ denotes a belief reward which is the expected reward of taking action $a$ at belief $b$. %

\subsection{Unknown Environment Coverage Problems}

\phdone{Coverage Problem}
For our coverage planning problem, we define the state as $s = (q, W)$, where $q$ is the robot state and $W$ is the world state. We maintain two representations of the world, i.e., $W = (W_{r}, W_{c})$, where $W_{r}$ denotes the world traversal risk state and $W_{c}$ is the world coverage state.

$W_{r}$ encodes the traversability risk of the world with respect to a robot's dynamic constraints. This state is critical in capturing traversability-stressing elements of the environment (slopes, rough terrain, and narrow passages, etc.) and is typically constructed by aggregating long-range sensor measurements. %
The cost function $C(W_{r}, q, a)$ returns the actuation effort and risk associated with executing action $a$ at robot state $q$ on $W_{r}$.

$W_{c}$ provides an estimation of what parts of the world have been observed, or \textit{covered}, by a particular sensor.
The coverage state is generated by specific sensor measurements, which may not necessarily be useful as navigational feedback, but instead are based on a task at hand. For instance, the coverage sensor may be a thermal camera for detecting thermal signatures, or a vision-based camera for identifying visual clues in the environment. 
As a robot moves, the sensor footprint sweeps the environment, expanding the covered area, or more generally, the task-relevant information about the world.

The coverage planning objective is to determine a trajectory through an environment that maximizes information gain $I$ while simultaneously minimizing action cost $C$. As such, the traversal risk and coverage states form the basis of the coverage reward function:
\begin{align}
  R(s, a) = f(I(W_{c}, a),\; C(W_{r}, q, a)),
  \label{eq:coverage_reward}
\end{align}
where $I(W_{c}, a) = H(W_{c}) - H(W_{c} \,|\, a)$ is quantified as reduction of the entropy $H$ in $W_{c}$ after taking action $a$. 
\phdone{Receding Horizon Planning}
Note that in unknown space coverage domains, we do not have strong priors about the parts of the world that have not yet been observed. Hence, knowledge about $W_{c}$ and $W_{r}$ in Eq.~(\ref{eq:coverage_reward}) at runtime is incomplete and often inaccurate.
Thus, in such domains, a Receding Horizon Planning (RHP) scheme has been widely adopted as the state-of-the-art \cite{bircher2016receding}.

\phdone{RHP Objective Function}
In POMDP formulation with RHP, the objective function in Eq.~(\ref{eq:objective_function}) is modified:
\begin{align}
  \pi_{t:t+T}^*(b) &= \argmax_{\pi \in \Pi_{t:t+T}} \, \mathbb{E} \sum_{t'=t}^{t+T} \gamma^{t'-t} r(b_{t'}, \pi_{t'}(b_{t'})),
  \label{eq:receding_objective_function}
\end{align}
where $T$ is a finite planning horizon for a planning episode at time $t$.
Given the policy from the last planning episode, only a part of the optimal policy, $\pi^*_{t:t+\Delta t}$ for $\Delta t \in (0, T]$, will be executed at runtime. A new planning episode will start at time $t+\Delta t$ with updated belief about $q$, $W_{c}$, and $W_{r}$.

\subsection{Challenges} \label{ssec:challenges}

We broadly identify the challenges associated with solving the unknown coverage planning problem, Eq.~(\ref{eq:receding_objective_function}), as computational complexity---in both time and space---and conflicting policy objectives over consecutive planning episodes, arising from unexpected updates in the belief at runtime.

\subsubsection{Time Complexity} \hfill

\noindent
POMDP planning suffers from \textit{the curse of dimensionality} \cite{KLC98} and \textit{the curse of history} \cite{Pineau03}. The former difficulty refers to fact that size of the belief grows exponentially with the size of the underlying state space. The latter difficulty refers to the fact that the number of action-observation sequences grows exponentially with the planning depth $d$, i.e., $\mathcal{O}(|\mathbb{A}|^d|\mathbb{Z}|^d)$.
As an example, for large-scale exploration of a 1~km-long environment with an action resolution of 1~m, 
the planning depth $d$ must be at least $10^3$ in order to reason about the coverage plan across the environment.

\subsubsection{Space Complexity} \hfill

\noindent
In addition to the classic time complexity of POMDPs, space complexity also poses a considerable challenge when handling the unknown environment coverage problem.
Since $s = (q, W)$, the space complexity is dominated by the world state $W$. 
For example, in a grid world, the memory complexity is $\mathcal{O}(|n|^k)$, with $n$ and $k$ denoting the number of discretization levels and the grid dimension, respectively. For a 1~km$^2$ environment at a 0.1~m resolution, %
with floating-point risk and coverage values stored in every cell, required memory is 800~MB. This amount of memory should be allocated for every search node, and thus the full space complexity of planning is $\mathcal{O}(|\mathbb{A}|^d |\mathbb{Z}|^d |n|^k)$ during each planning episode.

\subsubsection{Unexpected Belief Updates} \hfill

\noindent
As the robot explores its environment, it receives new sensory information, updates its belief, and constructs a new coverage policy in a receding horizon fashion.
Policies generated during consecutive planning episodes must respect the kinodynamic constraints of the robot, while simultaneously adapting to unexpected hazards in the environment.
We refer to these two distinct, and often opposed, objectives as \textit{consistency} and \textit{resiliency} of the receding-horizon policy, respectively.
Path consistency ensures smooth trajectories and continuous velocities during transitions from one policy to the next, while path resiliency ensures the path adapts to unexpected changes in the world risk state. %
Thus, it is imperative to find a balance between policy consistency and resiliency, particularly for safety-critical systems.

\section{PLGRIM: Hierarchical Coverage Planning on Information Roadmaps}
\label{sec:plgrim}

\phdone{Framework Overview}
In this section, we present a novel and field-hardened coverage planning autonomy framework, \textit{PLGRIM (Probabilistic Local and Global Reasoning on Information roadMaps)}, for exploration of large-scale unknown environments with complex terrain.
Our proposed methods to tackle the challenges described in Section~\ref{ssec:challenges} are:
\begin{enumerate}[label={\arabic*)}]
  \item \label{en:idea1} Space Complexity:
  We introduce a hierarchical belief space representation that is compact, versatile, and scalable.
  We refer to this representation as an Information RoadMap (IRM).
  Hierarchical IRMs can effectively encode a large-scale world state, while simultaneously capturing high-fidelity information locally.
  \item \label{en:idea2} Time Complexity:
  We propose hierarchical POMDP solvers that reason over long horizons within a suitable replanning time with locally near-optimal performance. Higher-level policies guide lower-level policies, resulting in a cascaded decision process.
  \item \label{en:idea4} Unexpected Belief Updates:
  We introduce a receding-horizon policy reconciliation method that respects the robot's dynamic constraints while ensuring resiliency to unexpected observations.
\end{enumerate}

\begin{figure}[t!]
  \centering
  \includegraphics[width=\columnwidth]{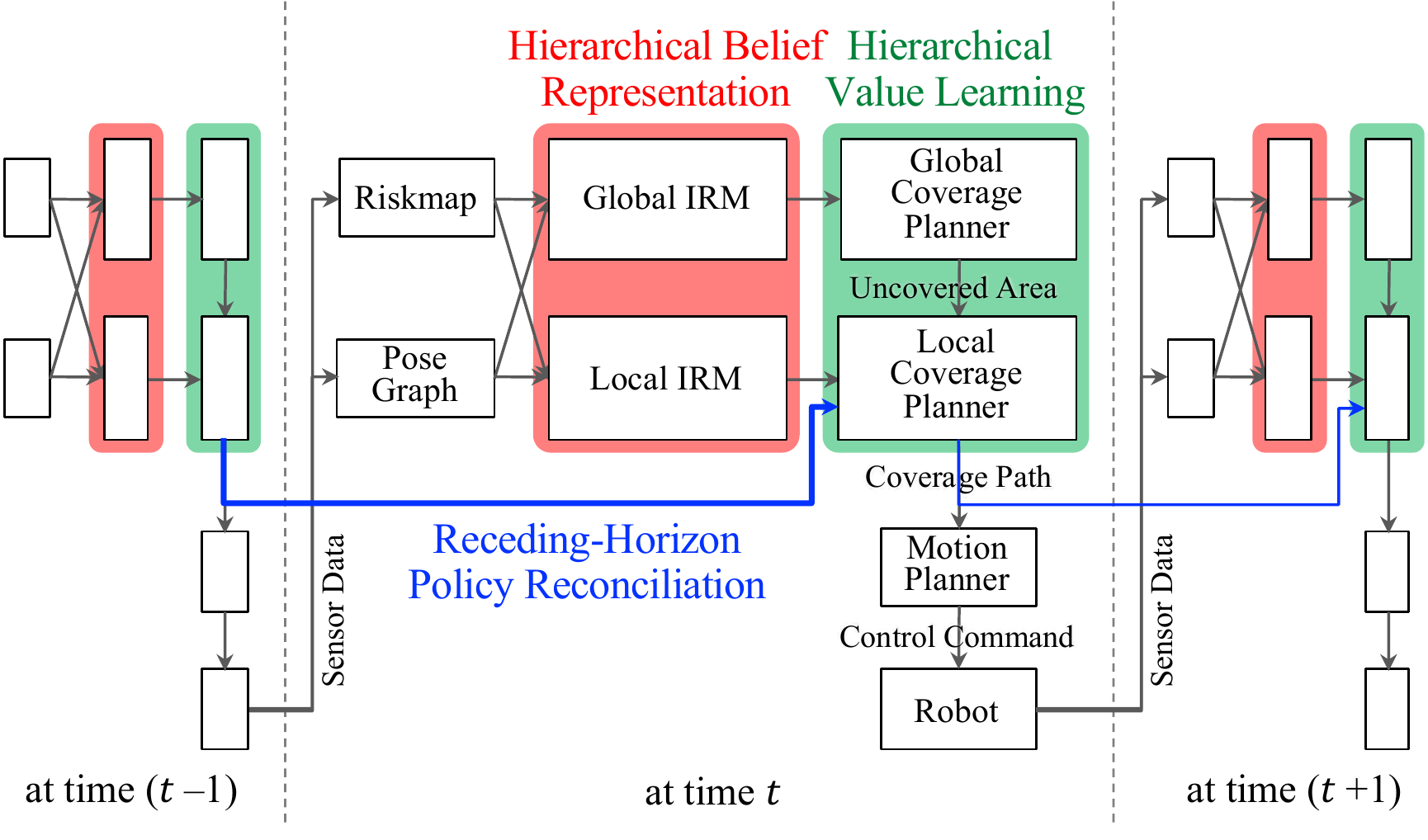}
  \caption{Illustration of PLGRIM framework for large-scale exploration in unknown environments.
  It \textit{i)} maintains hierarchical beliefs about the traversal risks and coverage states,
  \textit{ii)} performs hierarchical value learning to construct an exploration policy, and 
  \textit{iii)} reconciles policies over receding-horizon planning episodes.} 
  \label{fig:framework} 
\end{figure}

\noindent
In the following subsections, we provide the technical details about the proposed framework, illustrated in Fig.~\ref{fig:framework}.

\label{ssec:hierarchical_policy}

\subsection{Overview}
To enable efficient and reactive robot behaviors on very large scales, we decompose the problem into tractable subproblems by introducing spatial and temporal abstractions. Spatially, the belief space is approximated by a task-dependent structure, enriched with environment map estimates. Temporally, the belief space is approximated by the aggregation of multiple structures, each spanning a different spatial range. Finally, we introduce a cascaded optimization problem that returns a policy over the stratified belief space in real time.

\subsubsection{Belief Decomposition} \hfill

\noindent
Let us denote the global world state as $W^g$ and the local world state as $W^\ell$, which is a subset of the global state, i.e., $W^\ell \subset W^g$, around the robot.
We define local and global belief states as $b^\ell = p(q, W^\ell)$ and $b^g = p(q, W^g)$, respectively, where $p(W^\ell)$ is a local, robot-centric, rolling-window world belief representation with high-fidelity information, and $p(W^g)$ is a global, unbounded world belief representation with approximate information.

\subsubsection{Policy Decomposition} \hfill

\noindent
We decompose the policy into local and global policies: $\pi^\ell$ and $\pi^g$, respectively. The overall policy $\pi \in \Pi$ is constructed by combining the local and global policies:
\begin{align}
  &\pi(b) = \pi^\ell(b^\ell; \pi^g(b^g)).
\end{align}
We approximate the original RHP optimization problem in Eq.~(\ref{eq:receding_objective_function}) as cascaded hierarchical optimization problems as follows:
\begin{align}
  &\pi_{t:t+T}(b)
  = \argmax_{\pi \in \Pi_{t:t+T}} \, \mathbb{E} \sum_{t'=t}^{t+T} \gamma^{t'-t} r(b_{t'}, \pi(b_{t'}))
  \nonumber \\
  & \approx \argmax_{\pi^\ell \in \Pi^\ell_{t:t+T}} \, \mathbb{E} \sum_{t'=t}^{t+T} \gamma^{t'-t} r^\ell(b^\ell_{t'}, \pi^\ell(b^\ell_{t'}; \pi_{t:t+T}^g(b^g_t))),
  \label{eq:llp_optimization}
  \\
  &\text{where }
  \pi_{t:t+T}^g(b^g) = \argmax_{\pi^g \in \Pi^g_{t:t+T}} \, \mathbb{E} \sum_{t'=t}^{t+T} \gamma^{t'-t} r^g(b^g_{t'}, \pi^g(b^g_{t'})).
  \label{eq:glp_optimization}
\end{align}
\normalsize
where $r^\ell(b^\ell, \pi^\ell(b^\ell))$ and $r^g(b^g, \pi^g(b^g))$ are approximate belief reward functions for the local and global belief spaces, respectively.
Note that the codomain of the global policy $\pi^g(b^g)$ is a parameter space $\Theta^\ell$ of the local policy $\pi^\ell(b^\ell; \theta^\ell)$, $\theta^\ell \!\! \in \! \Theta^\ell\!$.\,

\phdone{Section Structure}
According to this formulation, we maintain the hierarchical belief representations (Section~\ref{ssec:belief-managers}) and solve for hierarchical POMDP policies (Section~\ref{ssec:belief-planners}).
For local planning consistency and resiliency, we extend Eq.~(\ref{eq:llp_optimization}) to a joint optimization problem given the previous planning episode policy (Section~\ref{ssec:resilient_rhp}).

\subsection{Hierarchical Belief Representation} \label{ssec:belief-managers}

\begin{algorithm}[t!]
{\small
\caption{Hierarchical IRM Construction}
\label{alg:IRMs}
\begin{algorithmic}
  \STATE \textbf{input:} Riskmap, Pose Graph %

  \STATE \textbf{\textit{\# Local IRM}}
  \STATE Local IRM $G^\ell = (N^\ell, E^\ell) \gets (\emptyset, \emptyset)$
  \STATE Add uniformly sampled nodes $\{n^\ell_i\}_i$ around the robot to $N^\ell$
  \FOR {each $n^\ell_i \in N^\ell$}
    \STATE Compute risk probability $p(n^\ell_{i,r})$ and coverage probability $p(n^\ell_{i,c})$ from Riskmap and Pose Graph for $n^\ell_i$
    \STATE Add $p(n^\ell_{i,r})$ and $p(n^\ell_{i,c})$ to the properties of $n^\ell_i$
  \ENDFOR
  \STATE Add edges for 8-connected neighbors, $\{e_{ij}\}^\ell_{i,j}$, to $E^\ell$
  \FOR {each $e^\ell_{ij} \in E^\ell$}
    \STATE Compute traversal risk $\rho_{ij}$ and distance $d_{ij}$ for $e^\ell_{ij}$
    \STATE Add $\rho_{ij}$ and $d_{ij}$ to the properties of $e^\ell_{ij}$
  \ENDFOR

  \STATE \textbf{\textit{\# Global IRM}}
  \IF {not initialized}
    \STATE Global IRM $G^g = (N^g_b \cup N^g_f, E^g) \gets (\emptyset, \emptyset)$
  \ENDIF
  
  \STATE Get the current robot pose $q$ from Pose Graph
  \IF {$q$ is farther from any breadcrumb node $\forall n^g_i \in N^g_b$ than $\bar{d}_b$}
    \STATE Add a new breadcrumb node $n^g = q$ to $N^g_b$
  \ENDIF

  \STATE Run \textsc{FrontierManager} to add new frontiers $\{n^g_{f^+}\}$ with coverage probabilities $\{p(n^g_{f^+,c})\}$, and prune invalidated frontiers, $\{n^g_{f^-}\}$, based on the current Riskmap and Pose Graph
  \STATE \hspace{7.2cm} $\triangleright$ \cite{keidar2012robot} %

  \FOR {each node $n^g_i \in \mathcal{N}_{G^g}(q)$}
    \FOR {each nearby node $n^g_j \in \mathcal{N}_{G^g}(n^g_i)$}
      \STATE Compute the traversal distance $d_{ij}$ and risk $\rho_{ij}$
      \IF {$d_{ij} < \bar{d_e}$ and $\rho_{ij} < \bar{\rho_e}$}
        \STATE Add an edge $e^g_{ij}$ to $E^g$ with properties $d_{ij}$ and $\rho_{ij}$
      \ELSE
        \STATE Remove the edge $e^g_{ij}$ from $E^g$
      \ENDIF
    \ENDFOR
  \ENDFOR

  \RETURN $G^\ell$ and $G^g$

\end{algorithmic}
} %
\end{algorithm}

\noindent
We introduce a hierarchical approximation of the belief space by decomposing the environment representation into multiple information-rich structures, each referred to as an Information Roadmap (IRM). 
We construct and maintain IRMs at two hierarchical levels: the \textit{Local IRM} and \textit{Global IRM}, as illustrated in Fig.~\ref{fig:IRMs}.

\subsubsection{World Belief Information Sources} \hfill

\noindent
During its exploration of an unknown environment, at any given time, the robot's understanding of the world is limited to noisy estimates of an observed subset of the world. 
IRMs are constructed from these estimates--namely, the \textit{Riskmap} and \textit{Pose Graph}.
A Riskmap, constructed through the aggregation of point cloud sensor measurements, is a local rolling-window map that provides risk assessment, 
effectively encoding the risk belief over the local world state $W^\ell$ \cite{fan2021step}.
A Pose Graph
estimates the past trajectory of the robot from relative pose measurements and informs the coverage belief over the global world state $W^g$ \cite{Ebadi2020}. 

\subsubsection{World Belief Construction} \hfill

\noindent
For compact and versatile representation of the world, we choose a generic graph structure, $G = (N, E)$ with nodes $N$ and edges $E$, as the data structure to represent the belief about the world state. Using this framework, nodes represent discrete areas in space, and edges represent actions. More precisely, we define an action as a motion control from the current node $n_i \in N$ to a neighboring node $n_j \in N$, connected by an edge $e_{ij} \in E$.

For a detailed description of the Local and Global IRM construction processes, see Algorithm~\ref{alg:IRMs}. We now describe the distinguishing features of each IRM: 

\begin{enumerate}[label={\arabic*)}]
  \itemsep0em 
  \setlength{\itemsep}{0.2em}
  \item \textit{Local IRM}: 
As an instantiation of the local world belief $p(W^\ell)$, we employ a rolling, fixed-sized grid structure $G^\ell=(N^\ell, E^\ell)$, which is centered at the robot's current position.
We uniformly sample nodes $n^\ell_i \in N^\ell$ from $W^\ell$,
and compute the risk and coverage probability distribution over a discrete patch centered at each node, i.e.,
$p(n^\ell_{i,r})$ and $p(n^\ell_{i,c})$, which are stored as node properties. 
For an edge $e^\ell_{ij}$, we compute and store the traversal distance $d_{ij}$ and risk $\rho_{ij}$,
which effectively encodes $p(W^\ell_{r})$ between two connected nodes.
In summary, the Local IRM contains relatively high-fidelity information at a high resolution, but locally.

  \item \textit{Global IRM}:
As an instantiation of the global world belief $p(W^g)$, we employ a sparse bidirectional graph structure $G^g=(N^g, E^g)$, which is fixed in the global reference frame.
Due to the space complexity concerns detailed in Section~\ref{ssec:challenges}, a densely-sampled grid structure, like $G^\ell$, is not a viable option for $G^g$, as it should span up to several kilometers.
Instead, we sparsely and non-uniformly sample nodes $n^g_i \in N^g$ from $W^g$ based on certain node-classifying conditions. Specifically, $N^g$ contains two mutually exclusive subsets of nodes: \textit{breadcrumbs} 
and \textit{frontiers}. 
Breadcrumb nodes are sampled directly from the Pose Graph, and thus capture the \textit{covered traversable} space of $W^g$.
Alternatively, frontier nodes are sampled from the border between covered and uncovered areas, and thus capture the \textit{uncovered traversable} space of $W^g$. 
Finally, in order for such a candidate node $n^g_i$ to be added to $G^g$, there must exist a traversable path to at least one nearby node $n^g_j \in N^g$. If such a path exists, an edge $e^g_{ij}$, storing traversal distance $d_{ij}$ and risk $\rho_{ij}$, is added to $G^g$. 
See Fig.~\ref{fig:graph-level-planner} for identification of breadcrumb and frontier nodes in $G^g$.
In summary, the Global IRM captures the free-space connectivity of $W^g$ with a notion of coverage, and does not explicitly encode highly-likely untraversable or uncertain areas in $W^g$ in order to achieve compact representation of the large-scale environment.

\end{enumerate}

\subsection{Hierarchical Value Learning} \label{ssec:belief-planners}

Given Local and Global IRMs as the hierarchical belief representation, we solve the cascaded hierarchical POMDP problems, Eq.~(\ref{eq:llp_optimization}) and Eq.~(\ref{eq:glp_optimization}), for coverage in an unknown environment.

\subsubsection{Solver Formulation} \hfill

\noindent
We start by introducing some notations.
We define \textit{value function} $V(b; \pi)$ as the expected reward of following policy $\pi$, starting from belief $b$:
\begin{align}
  V(b; \pi) &= \mathbb{E} \Big[ \sum_t \gamma^t r(b_t, \pi(b_t))] \Big].
\end{align}
From a recursive form of the value function, we can define the value of taking action $a$ in belief $b$ under a policy $\pi$ by the \textit{action-value function}:
\begin{align}
  Q(b, a; \pi) = r(b, a) + \sum_{b' \in \mathbb{B}} \gamma \, \mathcal{T}(b, a, b') \, V(b'; \pi),
  \label{eq:q_function}
\end{align}
where $\mathcal{T}(b, a, b')$ is the transition probability from $b$ to $b'$ by action $a$, as follows:
\begin{align}
  \mathcal{T}(b, a, b') = \sum_{z \in \mathbb{Z}} p(b' | b, a, z) \; p(z | b, a).
\end{align}
A POMDP solver tries to learn $Q(b, a)$ and $V(b) = \max_{a \in \mathbb{A}} Q(b, a)$, and returns the policy $\pi$ that specifies the best action for a given belief $b$, i.e., $\pi(b) = \argmax_{a \in \mathbb{A}} Q(b, a)$.

\subsubsection{Generalized Coverage Reward} \hfill

\noindent
Entropy provides a measure of uncertainty of a random variable's belief. Given an IRM $G = (N, E)$ containing a $p(n_{i,c})$ value for each node $n_i \in N$, the entropy of the world coverage state is:
\begin{multline}
  H(p(W_{c})) = - \sum_{i}^{|N|}\Big[ p(n_{i,c}) \log p(n_{i,c}) \\ 
  + p(n_{i,\neg c}) \log p(n_{i,\neg c}) \Big].
\end{multline}
If $a \in \mathbb{A}$ is a motion from node $n_i \in N$ to node $n_j \in N$ along edge $e_{ij} \in E$, then the coverage information gain (i.e., coverage uncertainty reduction) in \textit{coverage belief} $p(W_{c})$ induced by $a$ is defined as:
\begin{align}
    I(W_{c} \, | \, a) &= \underbrace{H(p(W_{c}))}_\text{current entropy} - \underbrace{H(p(W_{c}\, | \, a))}_\text{future entropy},
\end{align}
where the second term represents the expected future entropy of the world coverage state after execution of action $a$. 

Although the action cost function at each hierarchical level is dependent upon the IRM's particular action set $E$ (i.e., Local and Global IRMs have different action sets), it can be generically formulated as:
\begin{align}
    C(W_{r}, q, a) = k_d d_{ij} + k_\rho \rho_{ij} + k_\mu \mu_{ij}(q),
\end{align}
where $d_{ij}$ and $\rho_{ij}$ are the traversal distance and risk along edge $e_{ij}$, respectively. The cost $\mu_{ij}(q)$ is associated with the current motion primitive, and is a consequence of the robot's non-holonomic constraints, such as the heading direction. Constants $k_d$, $k_\rho$, and $k_\mu$ weigh the importance of traversal distance, risk, and motion primitive history on the total action cost.

Then, finally the coverage reward function is defined as a weighted sum of the information gain and action cost:
\begin{align}
    R(s, a) = k_I I(W_{c}, z) - k_C \, C(W_{r}, q, a)),
\end{align}
where $k_I$ and $k_C$ are constant weights.

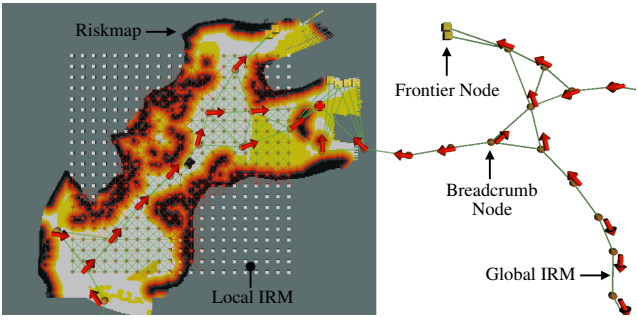
\begin{figure}[t!]
\centering
    \begin{tikzpicture}
    \node[anchor=south west,inner sep=0] (image) at (0,0) {\includegraphics[width=1\columnwidth]{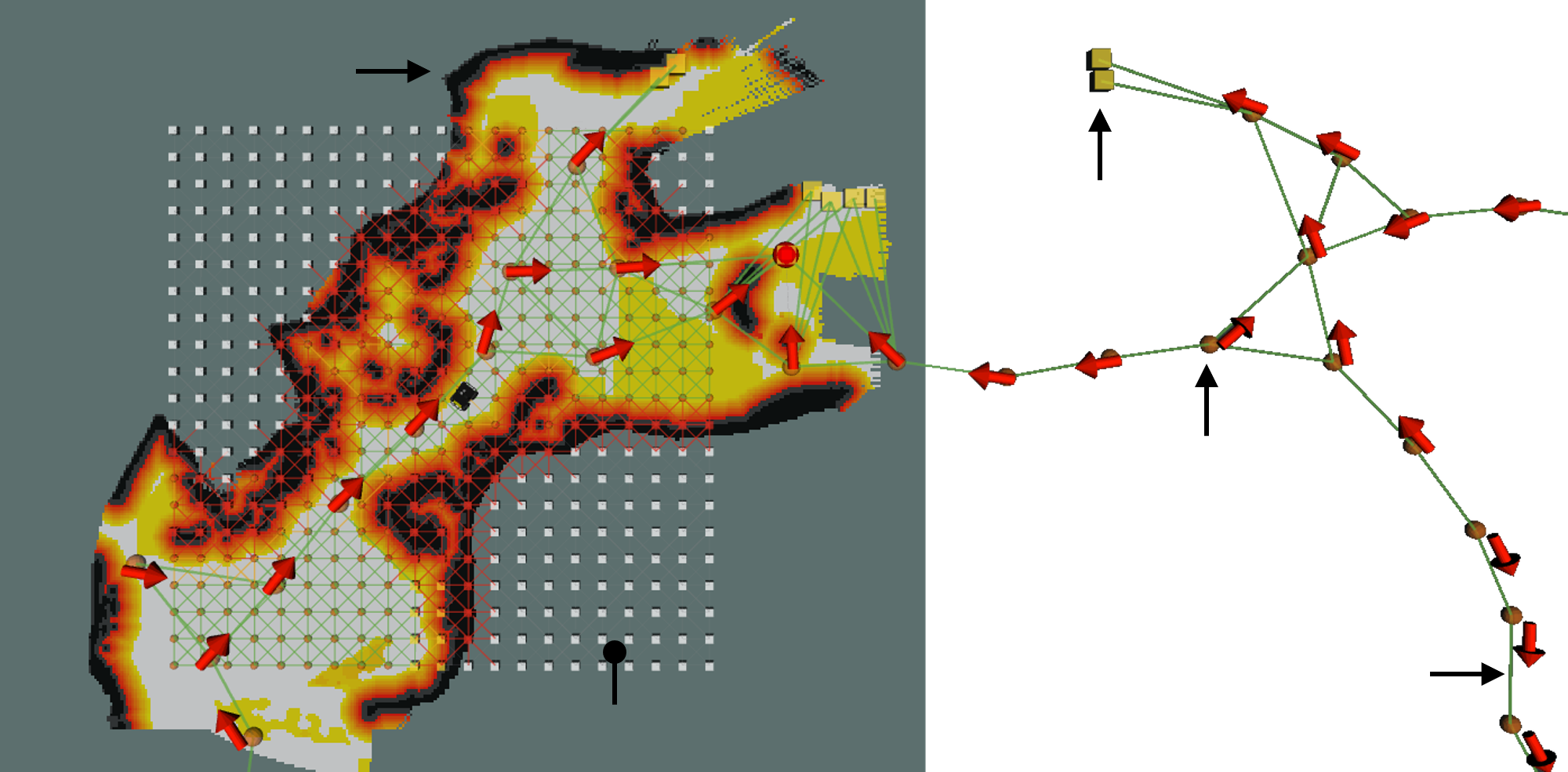}};
	    \begin{scope}[x={(image.south east)},y={(image.north west)}]

	    	\node [font=\scriptsize,above left,align=right,black] at (0.8,0.67) {Frontier Node}; 
	    	\node [font=\scriptsize,above left,align=right,black] at (0.86,0.35) {Breadcrumb}; 
	    	\node [font=\scriptsize,above left,align=right,black] at (0.82,0.2915) {Node}; 
	    	\node [font=\scriptsize,above left,align=right,black] at (0.235,0.85) {Riskmap}; 
	    	\node [font=\scriptsize,above left,align=right,black] at (0.475,0.008) {Local IRM}; 
	    	\node [font=\scriptsize,above left,align=right,black] at (0.92,0.081) {Global IRM}; 

	    \end{scope}
	\end{tikzpicture}	
  \caption{QMDP policy (red arrows displayed above \textit{breadcrumb} nodes) for Global Coverage Planning (GCP). A red sphere indicates the QMDP frontier goal.}
  \label{fig:graph-level-planner}
\end{figure}

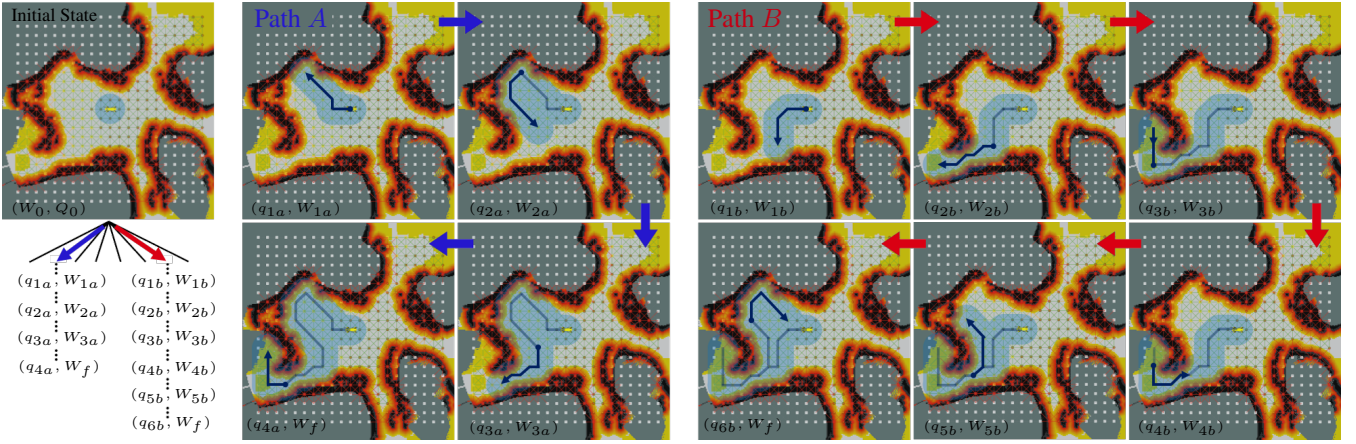
\begin{figure*}[h!]
\centering
    \begin{tikzpicture}
	    \node[anchor=south west,inner sep=0] (image) at (0,0) {\includegraphics[width=1\textwidth]{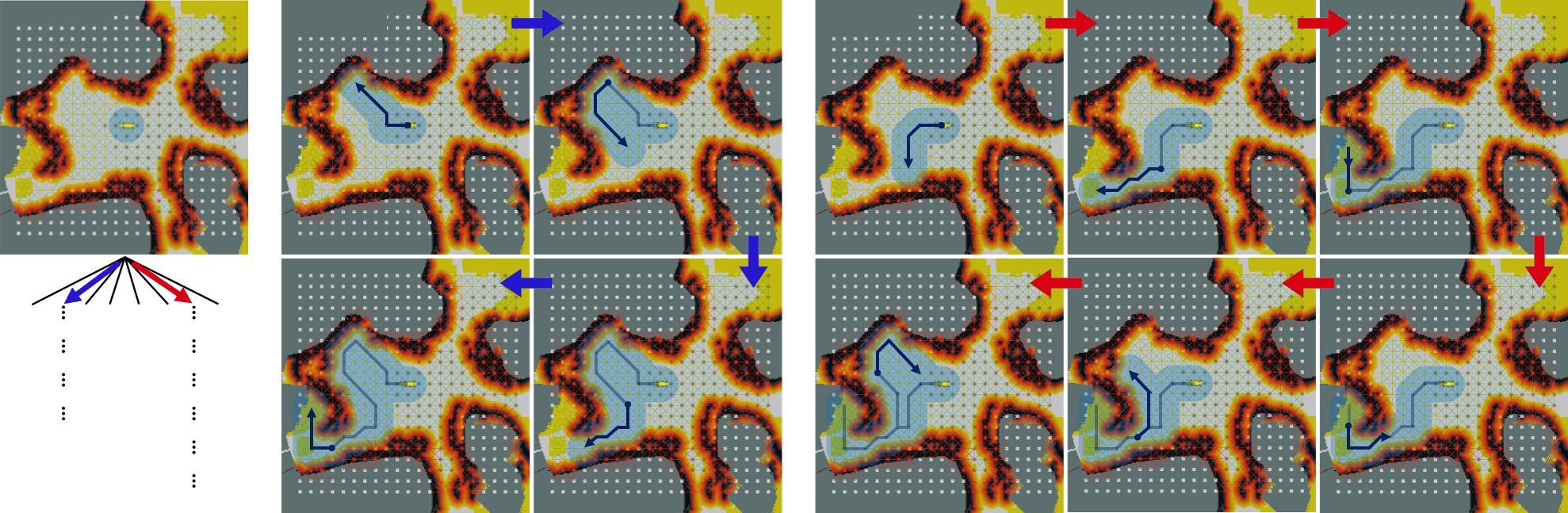}};
	    \begin{scope}[x={(image.south east)},y={(image.north west)}]

	        \definecolor{mycolor_red}{RGB}{195.2, 0.0, 19.8}
	        \definecolor{mycolor_blue}{RGB}{37.2, 24, 204.6}
	        \node [above left,align=right,color=mycolor_blue] at (0.25,0.92) {Path $A$};
	        \node [above left,align=right,color=mycolor_red] at (0.59,0.92) {Path $B$};
	        \node [font=\scriptsize, above left,align=right,black] at (0.078,0.935) {Initial State};

	    	\node [font=\fontsize{4pt}{0}, above left,align=right,black] at (0.07,0.49) {$(W_0,Q_0)$};

	    	\node [font=\fontsize{4pt}{0}, above left,align=right,black] at (0.0855,0.325) {$(q_{1a}, W_{1a})$};
	    	\node [font=\fontsize{4pt}{0}, above left,align=right,black] at (0.0855,0.26) {$(q_{2a}, W_{2a})$};
	    	\node [font=\fontsize{4pt}{0}, above left,align=right,black] at (0.0855,0.195) {$(q_{3a}, W_{3a})$};
	    	\node [font=\fontsize{4pt}{0}, above left,align=right,black] at (0.080,0.122) {$(q_{4a}, W_{f})$};
	    	
	    	\node [font=\fontsize{4pt}{0}, above left,align=right,black] at (0.1675,0.325) {$(q_{1b}, W_{1b})$};
	    	\node [font=\fontsize{4pt}{0}, above left,align=right,black] at (0.1675,0.26) {$(q_{2b}, W_{2b})$};
	    	\node [font=\fontsize{4pt}{0}, above left,align=right,black] at (0.1675,0.195) {$(q_{3b}, W_{3b})$};
	    	\node [font=\fontsize{4pt}{0}, above left,align=right,black] at (0.1675,0.1265) {$(q_{4b}, W_{4b})$};
	    	\node [font=\fontsize{4pt}{0}, above left,align=right,black] at (0.1675,0.063) {$(q_{5b}, W_{5b})$};
	    	\node [font=\fontsize{4pt}{0}, above left,align=right,black] at (0.163,-0.005) {$(q_{6b}, W_{f})$};

	    	\node [font=\fontsize{4pt}{0}, above left,align=right,black] at (0.26,0.49) {$(q_{1a}, W_{1a})$};
	    	\node [font=\fontsize{4pt}{0}, above left,align=right,black] at (0.423,0.49) {$(q_{2a}, W_{2a})$};
	    	\node [font=\fontsize{4pt}{0}, above left,align=right,black] at (0.6,0.49) {$(q_{1b}, W_{1b})$};
	    	\node [font=\fontsize{4pt}{0}, above left,align=right,black] at (0.76,0.49) {$(q_{2b}, W_{2b})$};
	    	\node [font=\fontsize{4pt}{0}, above left,align=right,black] at (0.92,0.49) {$(q_{3b}, W_{3b})$};

	    	\node [font=\fontsize{4pt}{0}, above left,align=right,black] at (0.25,-0.01) {$(q_{4a}, W_{f})$};
	    	\node [font=\fontsize{4pt}{0}, above left,align=right,black] at (0.423,-0.01) {$(q_{3a}, W_{3a})$};
	    	\node [font=\fontsize{4pt}{0}, above left,align=right,black] at (0.59,-0.01) {$(q_{6b}, W_{f})$};
	    	\node [font=\fontsize{4pt}{0}, above left,align=right,black] at (0.76,-0.01) {$(q_{5b}, W_{5b})$};
	    	\node [font=\fontsize{4pt}{0}, above left,align=right,black] at (0.92,-0.01) {$(q_{4b}, W_{4b})$};
	    \end{scope}
	\end{tikzpicture}	
  \caption{Illustrative example of coverage path planning on the Local IRM with Monte-Carlo Tree Search. The field-of-view of the robot's coverage sensor is represented by a blue circle. 
  Macro actions (6 steps on Local IRM in this example) associated with the two tree branches, paths \textit{A} and \textit{B}, are shown. Note that the final world coverage states in both branches are identical. Path \textit{A} is evaluated to be more rewarding than \textit{B} since fewer actions were required to cover the same area.}
  \label{fig:lattice-level-planner}
\end{figure*}

\subsubsection{Local-Global Coverage Planner Coordination} \hfill

\noindent
In our cascaded hierarchical optimization framework, we first solve for the global policy in Eq.~(\ref{eq:glp_optimization}). The global policy solution then serves as an input parameter to the local policy in Eq.~(\ref{eq:llp_optimization}). This means that Global Coverage Planner (GCP) provides \textit{global guidance} to the Local Coverage Planner (LCP).

The role of GCP is to construct a low-fidelity policy that provides global guidance to uncovered areas, at which point, LCP instructs a local coverage behavior. More concretely, a target frontier node in the Global IRM, $n^g_f \in N^g_f$, can be extracted from the global-level control $a^g \in \mathbb{A}^g$ provided by GCP. Since the environment can be very large ($>\!\!1$~km), GCP must be capable of reasoning over hundreds of nodes on the Global IRM. To alleviate this scalability challenge, we assume that GCP's policy terminates at frontier nodes. By classifying frontier nodes as terminal in the belief space, we can assume no changes occur to the world coverage state before termination. Therefore, we omit $W$ from the state space for GCP.

The role of LCP is to construct a high-fidelity policy that provides local guidance based on information gathering, traversal risk (e.g., proximity to obstacles, terrain roughness, and slopes), and the robot's mobility constraints (e.g., acceleration limits and non-holonomic constraints of wheeled robots).
LCP has two phases: i) reach the target area based on GCP's guidance, and ii) construct a local coverage path after reaching the target area.
If the target frontier is outside the Local IRM range, i.e., $n^g_f \notin W^\ell$, LCP simply instantiates high-fidelity control  based on the global-level control $a^g$ in order to reach the target frontier.
If the target frontier $n^g_f$ is within the Local IRM range, i.e., $n^g_f \in W^\ell$, then LCP performs the nominal information-gathering coverage optimization, as described in Eq.~(\ref{eq:llp_optimization}).

\subsubsection{Global Coverage Planner (GCP) Algorithm}\label{sssec:GCP} \hfill

\noindent
In this work, we adopt the QMDP approach for the global coverage planning problem \cite{littman1995learning}.
The key idea of QMDP is to assume the state becomes fully observable after one action, so that the \textit{value function} for further actions can be evaluated efficiently in an MDP (Markov Decision Process) setting.
In our global coverage planning domain, we define the first action to be the robot's relocation to a nearby node on the Global IRM. At this point, the robot pose is assumed to be fully observable, while the world risk and coverage states remain unchanged.

\phdone{QMDP Details}
More formally, we solve for $Q^g_{\mathrm{MDP}}(q^g, a^g)$ by ignoring uncertainty in the robot pose $q^g$ and changes in the world coverage state $W^g_{c}$.
In this MDP setting, $Q^g_{\mathrm{MDP}}(q^g, a^g)$ can be learned by Value Iteration very efficiently, even for long discount horizons.
Then, we evaluate the \textit{action-value function} in Eq.~(\ref{eq:q_function}) in a POMDP setting for the current belief and the feasible one-step actions:
\begin{align}
  Q(b^g, a^g) = \int_{q^g} b(q^g) Q^g_{\mathrm{MDP}}(q^g, a^g) \mathrm{d}q^g.
\end{align}
Finally, a POMDP policy can be obtained as follows:
\begin{align}
  \pi^g(b^g) = \argmax_{a^g \in \mathbb{A}^g} Q(b^g, a^g).
\end{align}
An example of the GCP policy is depicted in Fig.~\ref{fig:graph-level-planner}.

\subsubsection{Local Coverage Planner (LCP) Algorithm}\label{sssec:LCP} \hfill

\noindent
In order to solve Eq.~(\ref{eq:llp_optimization}), we employ POMCP (Partially Observable Monte Carlo Planning) algorithm \cite{silver2010monte}.
POMCP is a widely-adopted POMDP solver that leverages the Monte Carlo sampling technique to alleviate both of the \textit{curse of dimensionality} and \textit{history}.
Given a generative model (or a black box simulator) for discrete action and observation spaces, POMCP can learn the \textit{value function} of the reachable belief subspace with an adequate exploration-exploitation trade-off.

\phdone{POMCP Details}
More concretely, POMCP evaluates $Q^\ell(b^\ell, a^\ell)$ in Eq.~(\ref{eq:q_function}) by unrolling recursive value backpropagation through sampled action-observation sequences. 
UCT algorithm for action selection helps to balance between exploration and exploitation in order to learn the \textit{action-value function}.
Initially, it explores the search space (possible action-observation sequences) with a random or a heuristically guided rollout policy.
While incrementally building the belief tree, it gradually exploits the learned values for more focused exploration. 
See illustration of local coverage planning in Fig.~\ref{fig:lattice-level-planner}.

\subsection{Receding-Horizon Policy Reconciliation} \label{ssec:resilient_rhp}

\phdone{Consistency and Resiliency}
We extend the receding-horizon \textit{local} coverage planning problem to address the trade-off between policy consistency and resiliency, as described in Section~\ref{ssec:challenges}.

\phdone{Receding Horizon Planning Re-formulation}
We define a policy reconciliation optimization problem by introducing the previous planning episode policy into Eq.~(\ref{eq:receding_objective_function}) for the current planning episode.
For notational brevity, let us denote the time when the previous policy was generated as $t_0$ and the current time as $t_1 = t_0 + \Delta t$.
In order to reconcile consecutive policies over receding horizons, we extend Eq.~(\ref{eq:receding_objective_function}) as follows, given the previous policy $\pi_{t_0:t_0+T}^-$ constructed at time $t_0$ for a finite horizon of $T$:
\begin{multline}
  \pi_{t_1:t_1+T}^*(b; \pi_{t_0:t_0+T}^-) 
  \\
  = \argmax_{\pi \in \Pi_{t_1:t_1+T}} \, \left[ \mathbb{E} \sum_{t'=t_1}^{t_1+T} \gamma^{t'-t_1} r(b_{t'}, \pi(b_{t'})) \right.
  \\
   \left. - \lambda \mathcal{R}(\pi_{t_0:t_0+T}^-, \pi_{t_1:t_1+T}) \right],
  \label{eq:revised_receding_objective_function}
\end{multline}
where $\mathcal{R}(\pi_{t_0:t_0+T}^-, \pi_{t_1:t_1+T})$ is a regularizing cost function that penalizes inconsistency between the previous and current policies in terms of kinodynamic constraints, and $\lambda$ is a regularization weight parameter.
The first term in Eq.~(\ref{eq:revised_receding_objective_function}) pursues policy resiliency based on the up-to-date world belief, which may encode unexpected hazards, while the second term promotes policy consistency.

Since the conflict between policy consistency and resiliency is most severe
at the junction between two consecutive policies,
we decompose Eq.~(\ref{eq:revised_receding_objective_function}) into two time frames, $(t_1:t_1+\tau)$ and $(t_1+\tau:t_1+T)$ for $\tau \in [0,\, T-\Delta t]$, and formulate it as a simplified joint optimization problem for $\tau^*$ and $\pi_{t_1+\tau^*:t_1+T}^*$:
\begin{align}
  & \tau^* = \argmax_{\tau \in [0,\, T-\Delta t]} \mathbb{E} \sum_{t'=t_1}^{t_1+\tau} \gamma^{t'-t_1} r(b_{t'}, \pi_{t_0:t_0+T}^-(b_{t'})),
  \label{eq:resiliency_tau}
  \\
  & \pi_{t_1+\tau^*:t_1+T}^* \nonumber \\
  & \quad= \argmax_{\pi \in \Pi_{t_1+\tau^*:t_1+T}} \, \mathbb{E} \sum_{t'=t_1+\tau^*}^{t_1+T} \gamma^{t'-t_1} r(b_{t'}, \pi(b_{t'})),
  \label{eq:resiliency_pi}
  \\
  & \pi_{t_1:t_1+T}^* = [\pi_{t_1:t_1+\tau^*}^-; \, \pi_{t_1+\tau^*:t+T}^*].
  \label{eq:concatenated_pi}
\end{align}
\normalsize
\noindent
Policy reconciliation is performed in Eq.~(\ref{eq:resiliency_tau}) 
over a single optimization variable $\tau$. By re-evaluating the previous policy $\pi_{t_0:t_0+T}^-$ with updated robot-world belief $b_{t'}$, $\tau$ dictates how much of the new $\pi_{t_1:t_1+T}^*$ should be in agreement with the previous policy. 
Effectively, a larger $\tau$ promotes policy consistency, while a smaller $\tau$ promotes policy resiliency.

Given $\tau^*$ from Eq.~(\ref{eq:resiliency_tau}), the optimization problem in Eq.~(\ref{eq:resiliency_pi}) becomes identical to Eq.~(\ref{eq:receding_objective_function}), except the change of start time,
and can be solved by LCP, as described in Section~\ref{sssec:LCP}.  %
The final receding-horizon policy $\pi_{t_1:t_1+T}^*$ is then constructed by concatenating the previous policy and a new partial policy, as in Eq.~(\ref{eq:resiliency_pi}).

\section{Experimental Results}\label{sec:exp_results}
In order to evaluate our proposed framework, we perform high-fidelity simulation studies with a four-wheeled vehicle (Husky robot) and real-world experiments with a quadruped (Boston Dynamics Spot robot). Both robots are equipped with custom sensing and computing systems, enabling high levels of autonomy and communication capabilities~\cite{Otsu2020,AliNeBula21arXiv}. The entire autonomy stack runs in real-time on an Intel Core i7 processor with 32 GB of RAM. The stack relies on a multi-sensor fusion framework. The core of this framework is 3D point cloud data provided by LiDAR range sensors mounted on the robots~\cite{Ebadi2020}. We refer to our autonomy stack-integrated Spot as Au-Spot~\cite{AutoSpot}.

\subsection{Baseline Algorithms}
We compare our PLGRIM framework against a local coverage planner baseline (next-best-view method) and a global coverage planner baseline (frontier-based method).
\begin{enumerate}[label={\arabic*)}]
  \item \textit{Next-Best-View (NBV):}
	NBV first samples viewpoints in a neighborhood of the robot, and then plans a deterministic path over a high-fidelity local world representation to each viewpoint~\cite{bircher2016receding}. The set of viewpoint paths serves as the policy search space. Each policy in the space is evaluated, and NBV selects the policy with the maximum reward, computed using action cost and information gain from the world representation. While NBV is able to leverage local high-fidelity information, it suffers due to its spatially limited world belief and sparse policy space.
  \item \textit{Hierarchical Frontier-based Exploration (HFE)}:
	Frontier based exploration methods construct a global, but low-fidelity, representation of the world, where frontiers encode approximate local information gain. The set of frontiers serves as the policy search space. Exploration interleaves a one-step look-ahead frontier selection and the creation of new frontiers, until all frontiers have been explored. Hierarchical approaches can enhance the performance of frontier-based methods by modulating the spatial scope of frontier selection~\cite{umari2017autonomous}. However, while HFE is able to reason across the global world belief, it suffers from downsampling artifacts and a sparse policy space composed of large action steps.
\end{enumerate}
Note that in order to achieve reasonable performance in the complex simulated environments, we allow each baseline to leverage our Local and Global IRM structures as the underlying search space.

\begin{figure}[t!]
\centering
    \begin{tikzpicture}
	    \node[anchor=south west,inner sep=0] (image) at (0,0) {\includegraphics[width=0.85\columnwidth]{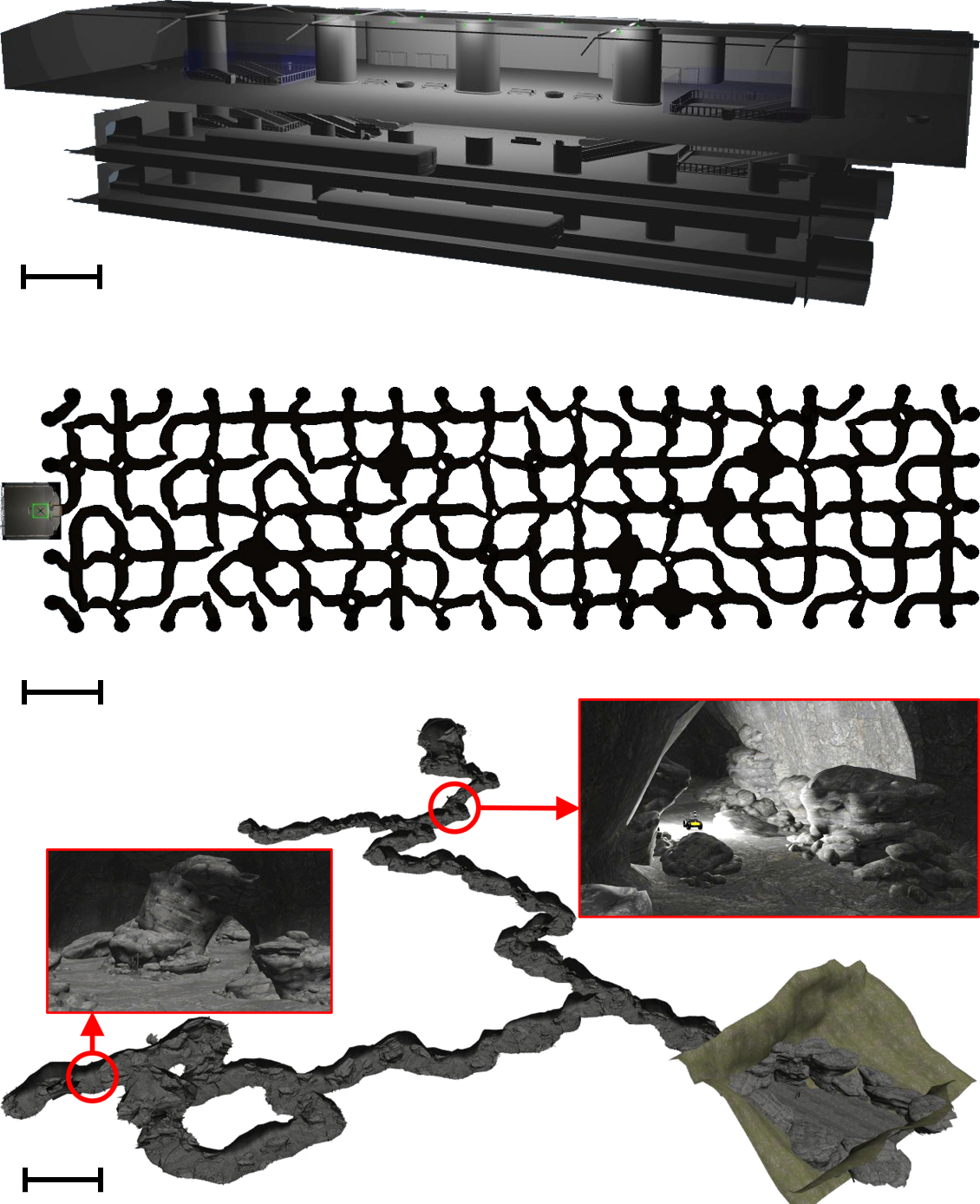}};
	    \begin{scope}[x={(image.south east)},y={(image.north west)}]

    	    \node [above left,align=right,black] at (0.475,0.99) {(a) Simulated Subway 1x};
        	\node [above left,align=right,black] at (0.37,0.67) {(b) Simulated Maze};
	    	\node [above left,align=right,black] at (0.37,0.33) {(c) Simulated Cave};

	    	\node [font=\scriptsize,above left,align=right,black] at (0.115,0.77) {10 m};
	    	\node [font=\scriptsize,above left,align=right,black] at (0.115,0.425) {50 m};
	    	\node [font=\scriptsize,above left,align=right,black] at (0.115,0.0225) {40 m};

	    \end{scope}
	\end{tikzpicture}	
\caption{PLGRIM's performance was evaluated in various simulated environments: (a) subway station, (b) maze (top-down view), and (c) cave.}
\label{fig:maps_of_cave}
\end{figure}

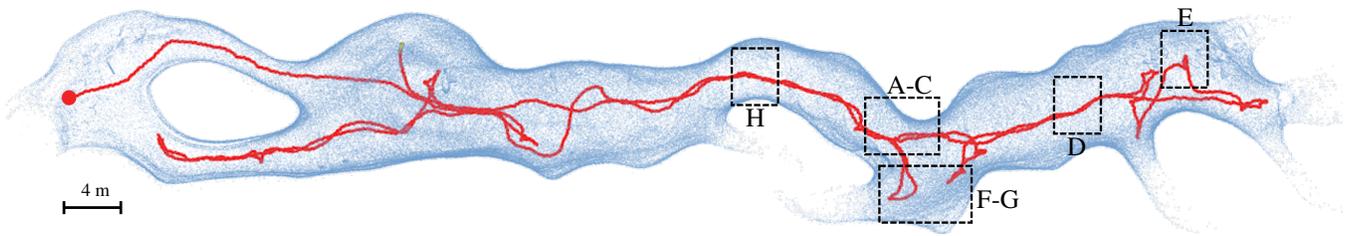
\begin{figure*}[h!]
\centering
    \begin{tikzpicture}
	    \node[anchor=south west,inner sep=0] (image) at (0,0) {\includegraphics[width=1\textwidth]{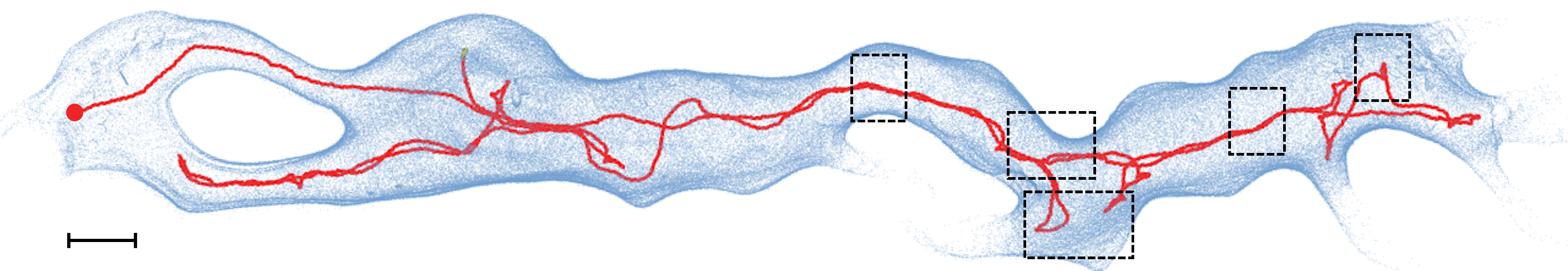}};
	    \begin{scope}[x={(image.south east)},y={(image.north west)}]

	    	\node [above left,align=right,black] at (0.575,0.422) {H};
	    	\node [above left,align=right,black] at (0.7,0.56) {A-C};
	    	\node [above left,align=right,black] at (0.765,0.07) {F-G};
	    	\node [above left,align=right,black] at (0.815,0.295) {D};
	    	\node [above left,align=right,black] at (0.895,0.855) {E};
	    	
	    	\node [font=\scriptsize, above left,align=right,black] at (0.0855,0.12) {4 m};
	    	
	    \end{scope}
	\end{tikzpicture}
	\caption{PLGRIM's exploration trajectory in Valentine Cave, Lava Beds National Monument, Tulelake, CA. Exploration started at the mouth of the cave (red circle), reached the end of the cave on the right, and returned back to visit uncovered areas. Boxes indicate the portions of the trajectory associated with the alphabetized snapshots in Fig.~\ref{fig:mlp_hardware_tests} and Fig.~\ref{fig:glp_hardware_tests}.
	}
    \label{fig:lava_tube_traj}
\end{figure*}

\begin{figure*}[h!]
\centering
    \begin{tikzpicture}
       \node[anchor=south west,inner sep=0] (image) at (0,0) {\includegraphics[width=1\textwidth]{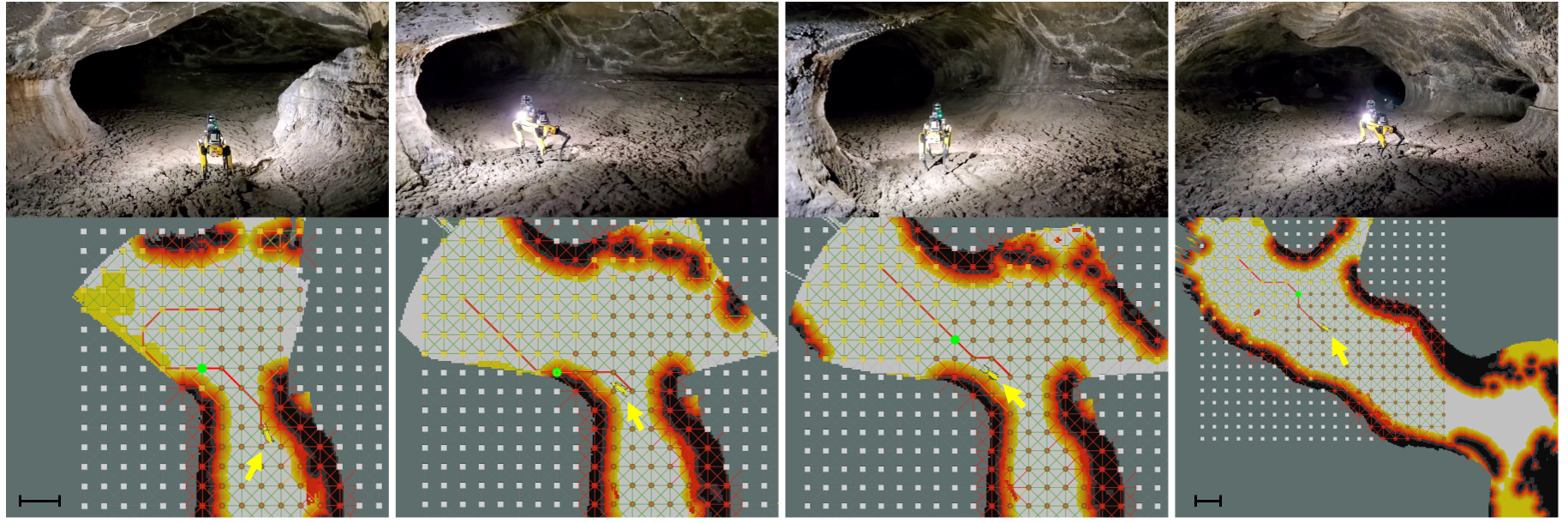}};
	    \begin{scope}[x={(image.south east)},y={(image.north west)}]

	    	\node [above left,align=right,white] at (0.035,0.91) {\textbf{A}};
	    	\node [above left,align=right,white] at (0.28,0.91) {\textbf{B}};
	    	\node [above left,align=right,white] at (0.53,0.91) {\textbf{C}};
	    	\node [above left,align=right,white] at (0.78,0.91) {\textbf{D}};

	    	\node [font=\scriptsize,above left,align=right,white] at (0.245,0.58) {\textbf{t=5:08}};
	    	\node [font=\scriptsize,above left,align=right,white] at (0.495,0.58) {\textbf{t=5:20}};
	    	\node [font=\scriptsize,above left,align=right,white] at (0.742,0.58) {\textbf{t=5:24}};
	    	\node [font=\scriptsize,above left,align=right,white] at (.992,0.58) {\textbf{t=6:24}};

	    	\node [font=\scriptsize,above left,align=right,black] at (0.044,0.045) {2 m};
	    	\node [font=\scriptsize,above left,align=right,black] at (0.787,0.045) {2 m};

	    \end{scope}
	\end{tikzpicture}	
\caption{The Local IRM (yellow, brown, and white nodes represent uncovered, covered and unknown areas, respectively) is shown overlaid on the Riskmap. A yellow arrow indicates the robot's location. LCP plans a path (red) that fully covers the local area (snapshot A). When $p(W^\ell)$ updates, the path is adjusted to extend towards the large uncovered swath while maintaining smoothness with the previous path. Another $p(W^\ell)$ update reveals that the path has entered a hazardous area---wall of lava tube (snapshot B). As a demonstration of LCP's resiliency, the path shifts away from the hazardous area, and the robot is re-directed towards the center of the tube (snapshot C). One minute later, the robot encounters a fork in the cave. The LCP path curves slightly toward fork apex (for maximal information gain) before entering the wider, less-risky channel (snapshot D). } 
\label{fig:mlp_hardware_tests} 
\end{figure*}
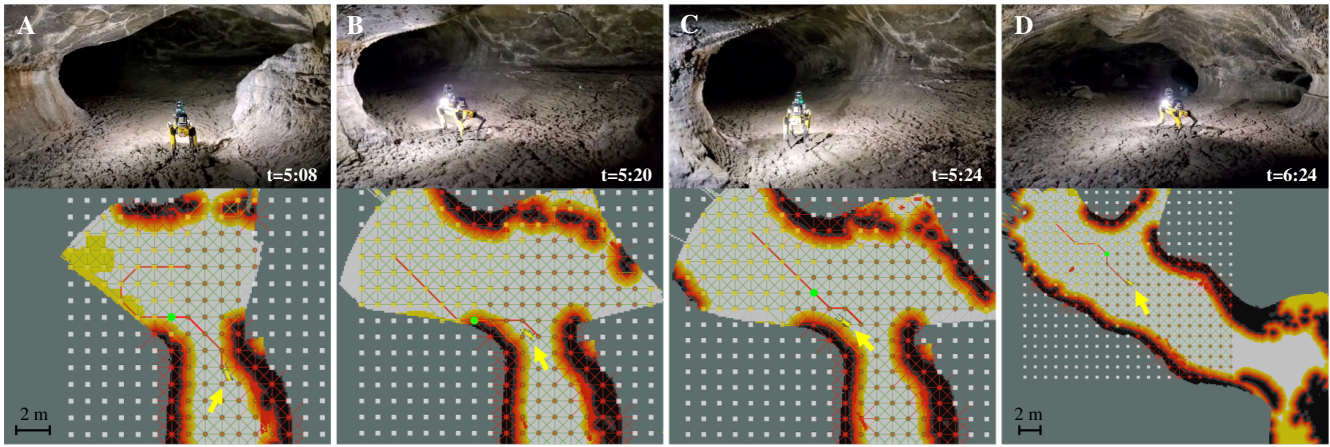

\begin{figure*}[h!]
\centering
    \begin{tikzpicture}
	    \node[anchor=south west,inner sep=0] (image) at (0,0) {\includegraphics[width=1\textwidth]{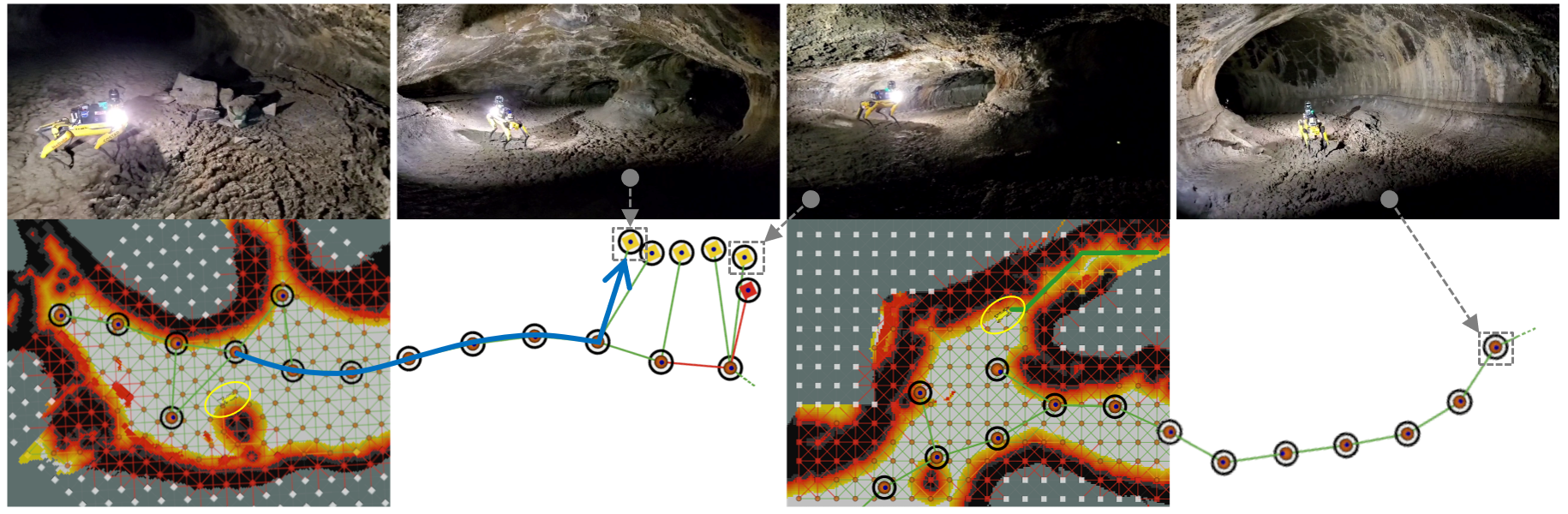}};
	    \begin{scope}[x={(image.south east)},y={(image.north west)}]

	    	\node [above left,align=right,white] at (0.035,0.91) {\textbf{E}};
	    	\node [above left,align=right,white] at (0.28,0.91) {\textbf{F}};
	    	\node [above left,align=right,white] at (0.53,0.91) {\textbf{G}};
	    	\node [above left,align=right,white] at (0.78,0.91) {\textbf{H}};

	    	\node [font=\scriptsize,above left,align=right,white] at (0.245,0.58) {\textbf{t=09:11}};
	    	\node [font=\scriptsize,above left,align=right,white] at (0.495,0.58) {\textbf{t=10:36}};
	    	\node [font=\scriptsize,above left,align=right,white] at (0.742,0.58) {\textbf{t=12:26}};
	    	\node [font=\scriptsize,above left,align=right,white] at (.992,0.58) {\textbf{t=14:16}};

	    \end{scope}
	\end{tikzpicture}	
\caption{Portions of the Global IRM constructed in the lava tube are visualized--yellow nodes represent frontiers, brown nodes represent breadcrumbs. Gray arrows associate a frontier with a snapshot of the robot exploring that frontier. GCP plans a path (blue) along the Global IRM to a target frontier after the local area is fully covered (snapshot E). The robot explores the area around the frontier (snapshot F), and then explores a neighboring frontier at the opening of a narrow channel to its right. LCP plans a path (green) into the channel (snapshot G). Later, after all local areas have been explored, the robot is guided back towards the mouth of cave along the breadcrumb nodes (snapshot H).} \label{fig:glp_hardware_tests} 
\end{figure*}
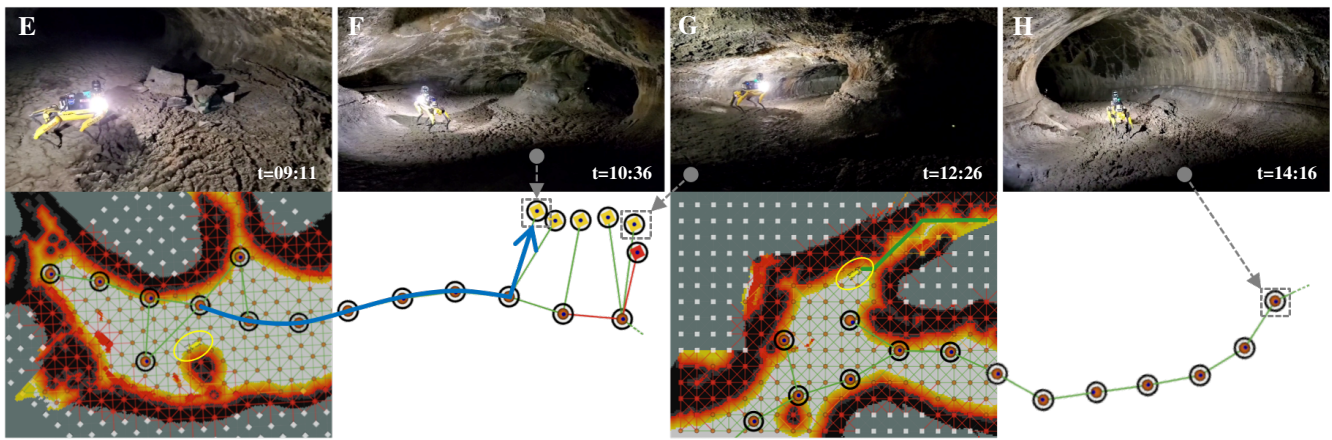

\subsection{Simulation Evaluation}
We demonstrate PLGRIM's performance, as well as that of the baseline algorithms, in a simulated subway, maze, and cave environment. Fig.~\ref{fig:maps_of_cave} visualizes these environments.

\subsubsection{Simulated Subway Station} \hfill

\begin{figure*}[t!]
    \centering
		\subfloat[][Simulated \\ \; \; \; Subway 1x]{\includegraphics[height=0.5759\columnwidth]{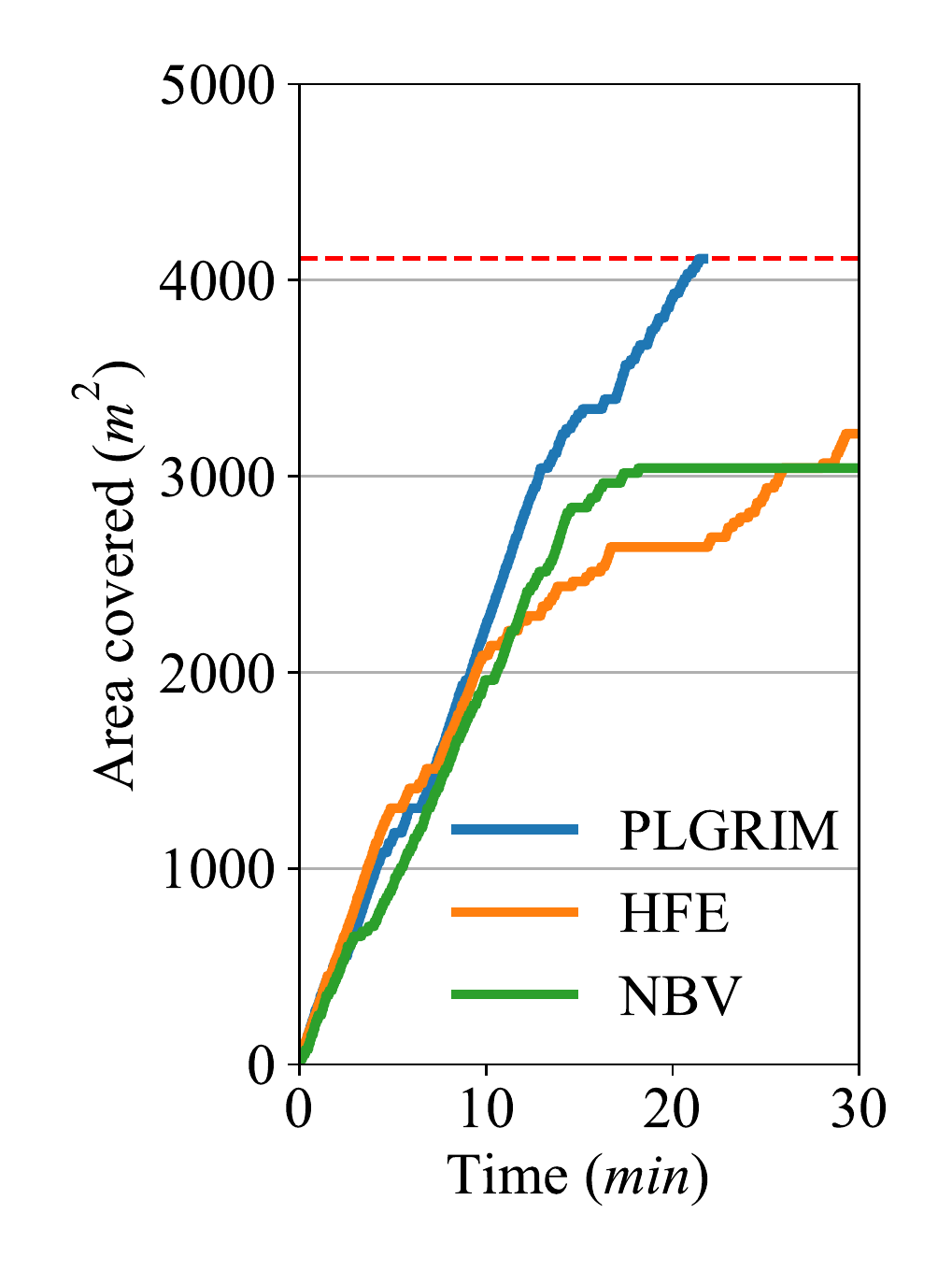}}
		\subfloat[][Simulated \\ \; \; \; Subway 2x]{\includegraphics[height=0.57015\columnwidth]{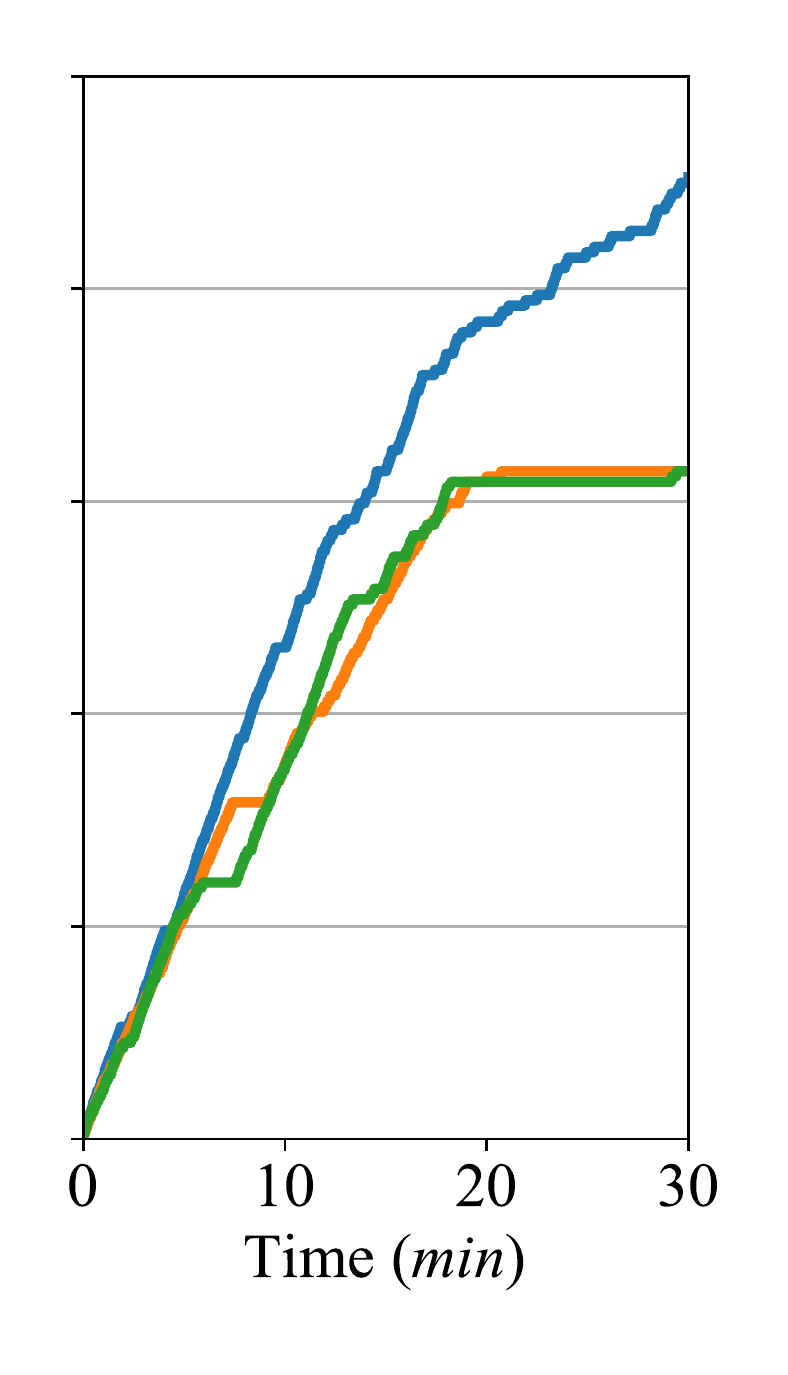}}	
		\subfloat[][Simulated \\ \; \; \; Subway 3x]{\includegraphics[height=0.57015\columnwidth]{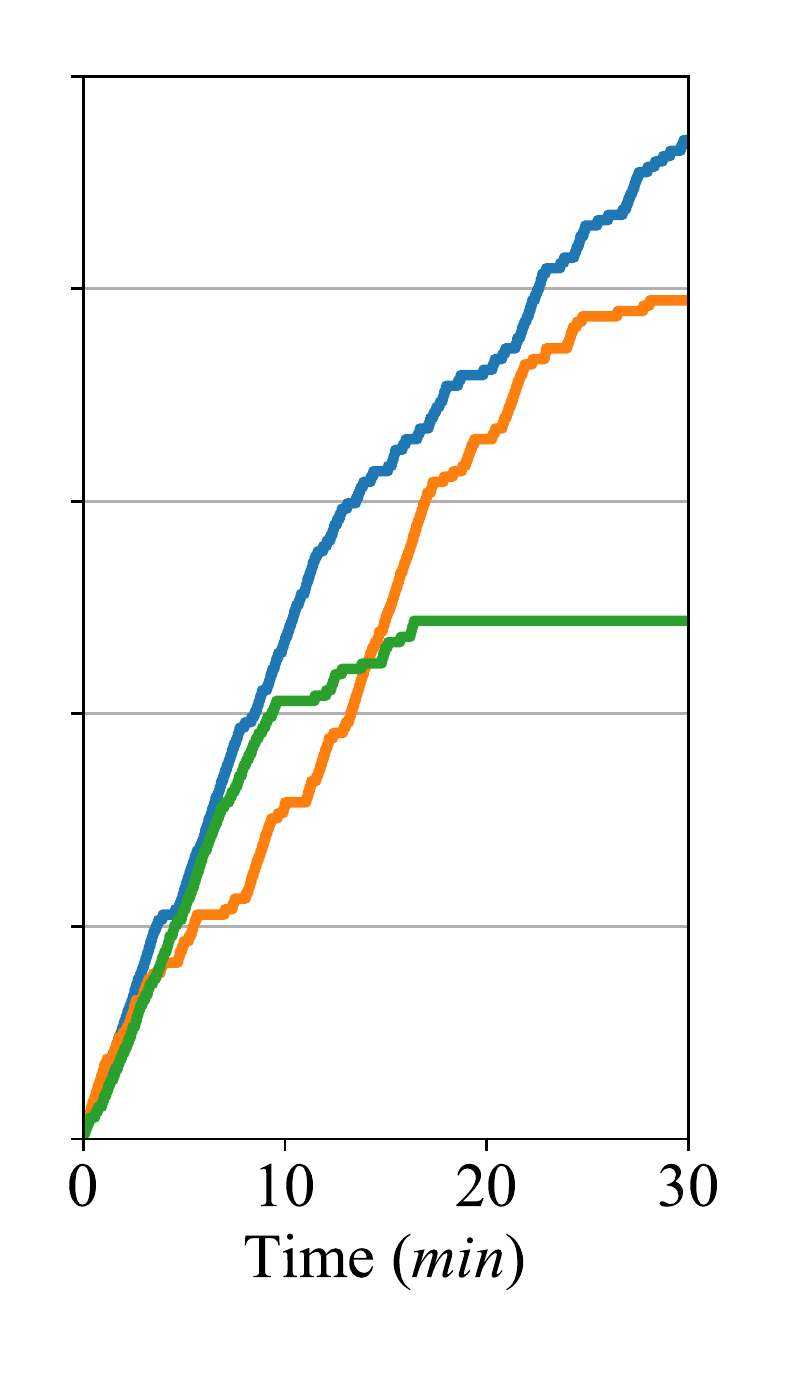}}	
        \subfloat[][Simulated \\Maze]{\includegraphics[height=0.57015\columnwidth]{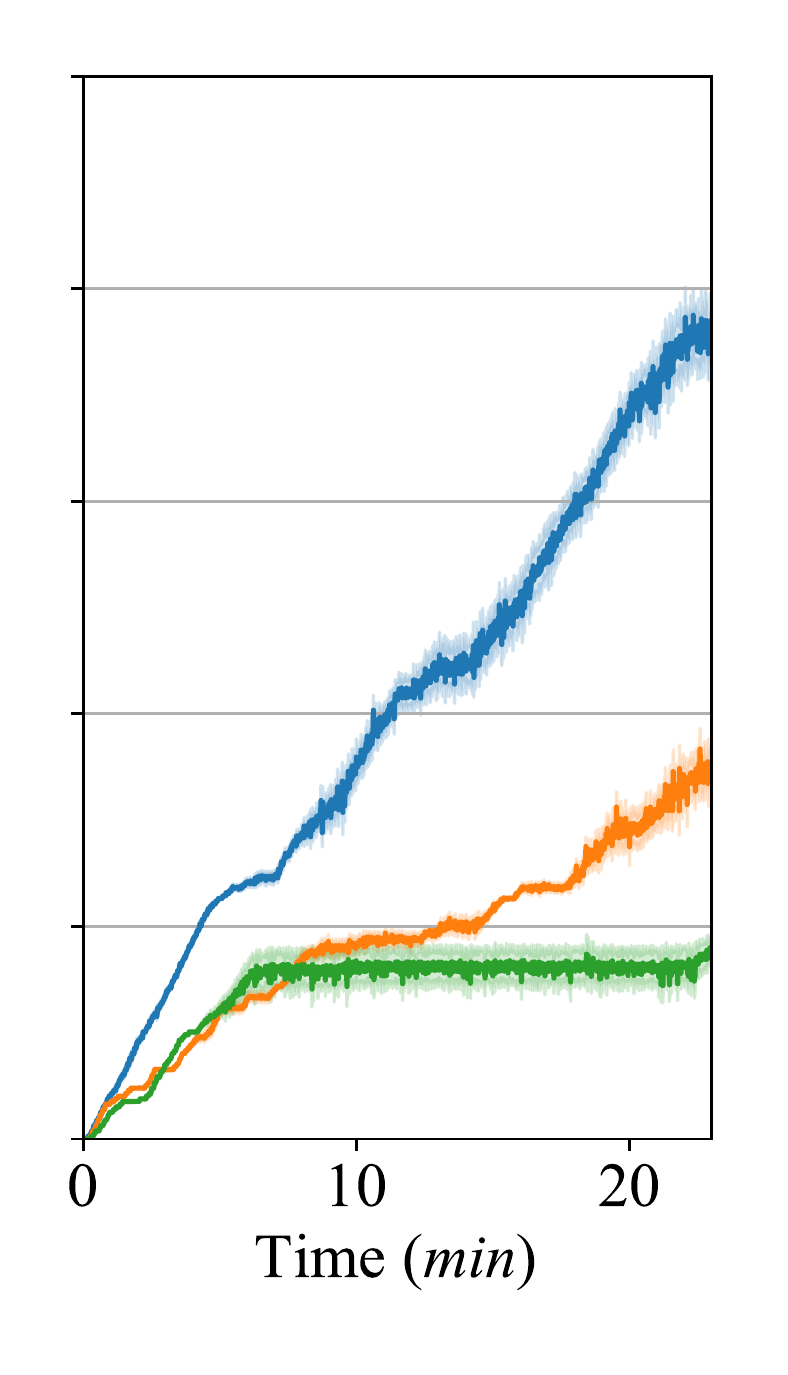}}
        \subfloat[][Simulated\\Cave]{\includegraphics[height=0.57015\columnwidth]{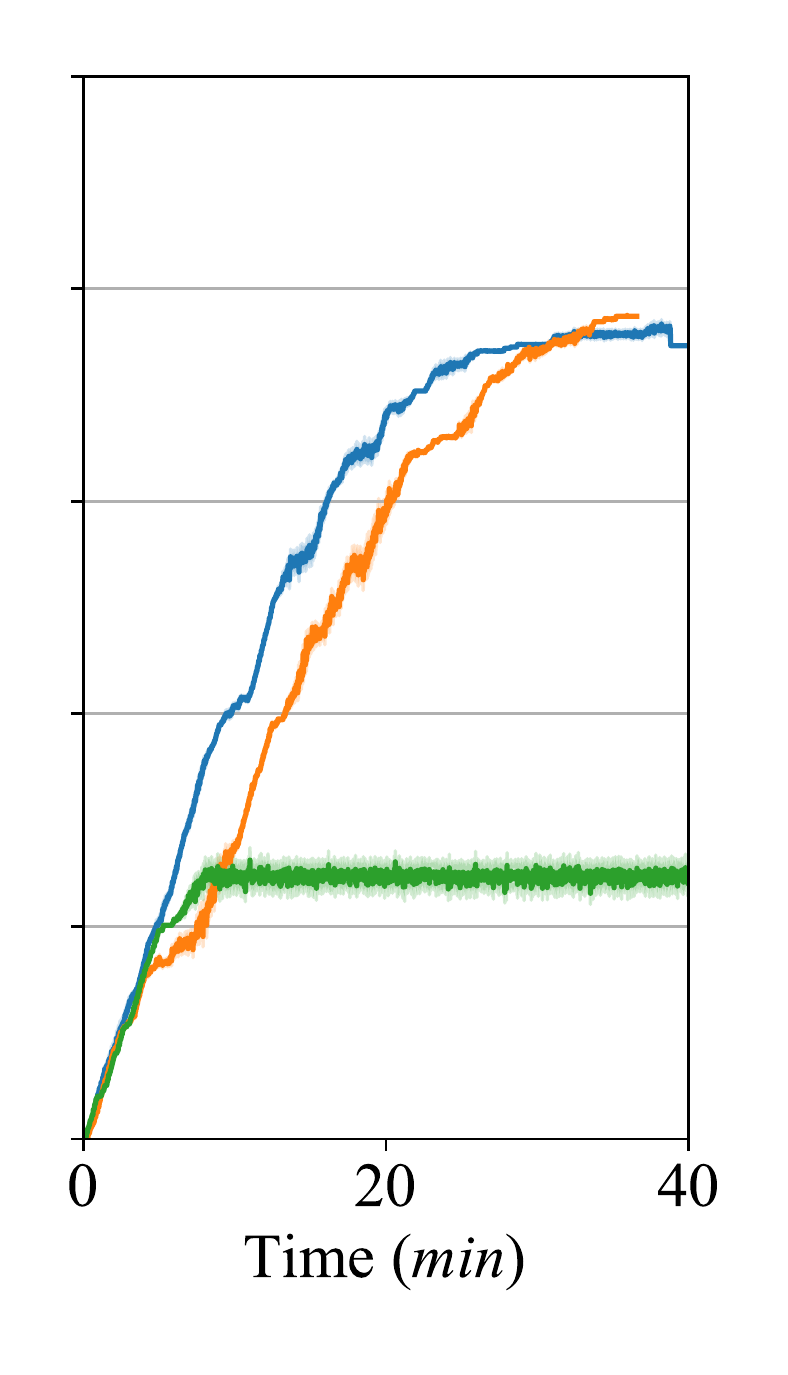}}
		\subfloat[][Real-world \\ \; \; Lava Tube]{\includegraphics[height=0.5759\columnwidth]{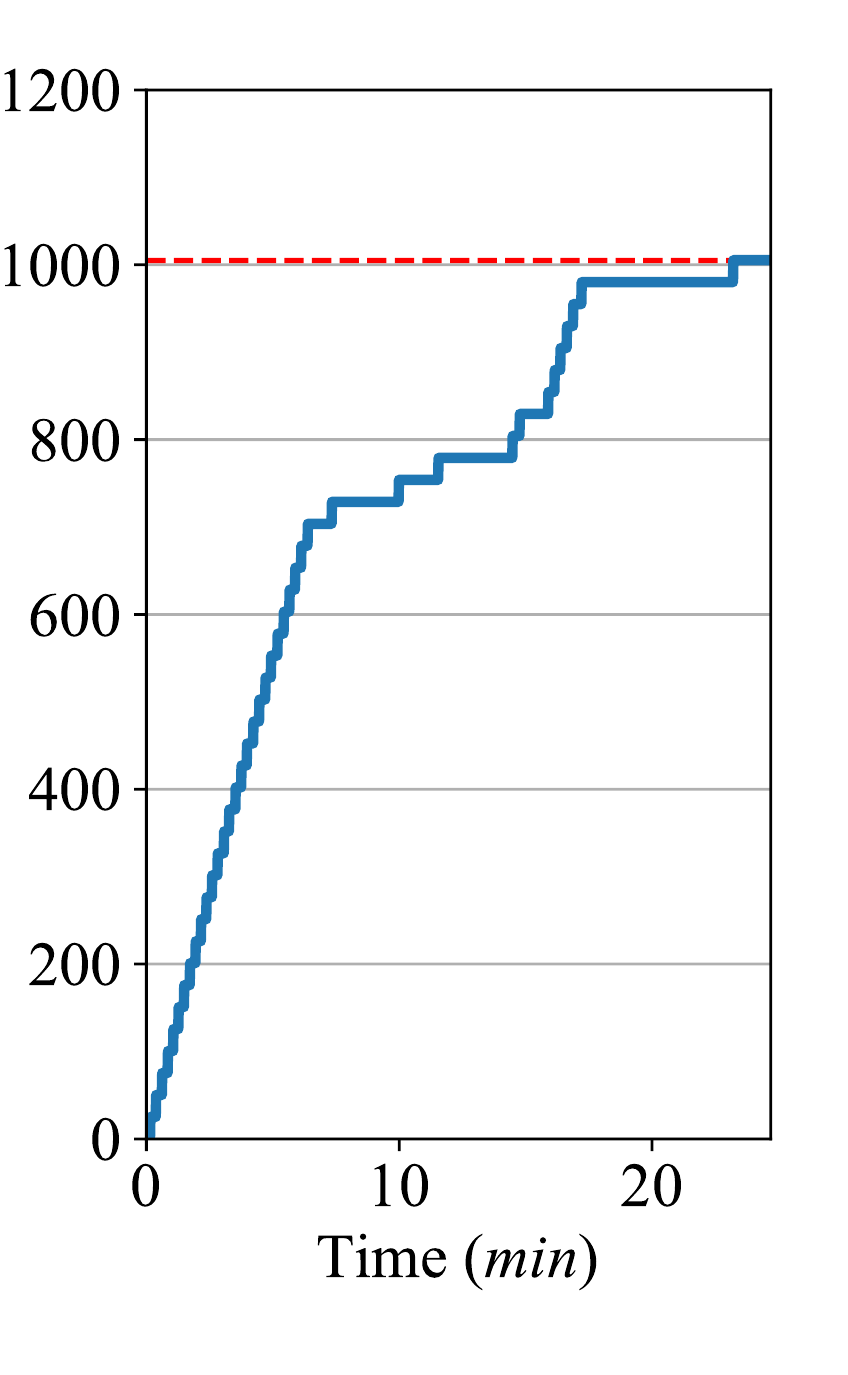}}
		\caption{Exploration by PLGRIM and baseline methods in simulated subway environments of increasing size (a)-(c), and in simulated and real-world cave environments (d)-(f). For (d) and (e), the covered area is the average of two runs. Red dashed lines indicate 100\% coverage of the environments, where applicable. 
		}
    \label{fig:all_together_plot}
\end{figure*}

\noindent
The subway station consists of large interconnected, polygonal rooms with smooth floors, devoid of obstacles. There are three varying sized subway environments, whose scales are denoted by 1x, 2x, and 3x. 
Fig.~\ref{fig:all_together_plot}(a)-(c) shows the scalable performance of PLGRIM against the baselines. 
In a relatively small environment without complex features (Subway 1x), NBV performance is competitive as it evaluates high-resolution paths based on information gain.
However, as the environment scale grows, its myopic planning easily gets \textit{stuck} and the robot's coverage rate drops significantly. 
HFE shows inconsistent performance in the subway environments. The accumulation of locally suboptimal decisions, due to its sparse environment representation, leads to the construction of a globally inefficient IRM structure. As a result, the robot must perform time-consuming detours in order to \textit{pick up} leftover frontiers.

\subsubsection{Simulated Maze and Cave} \hfill

\noindent
The maze and cave are both unstructured environments with complex terrain (rocks, steep slopes, etc.) and topology (narrow passages, sharp bends, dead-ends, open-spaces, etc.).
The coverage rates for each algorithm are displayed in Fig.~\ref{fig:all_together_plot}(d)-(e). 
PLGRIM outperforms the baseline methods in these environments. By constructing long-horizon coverage paths over a high-resolution world belief representation,
PLGRIM enables the robot to safely explore through hazardous terrain. 
Simultaneously, it maintains an understanding of the global world, which is leveraged when deciding where to explore next after exhausting all local information.
In the cave, NBV's reliance on a deterministic path, without consideration of probabilistic risk, causes the robot to drive into a pile of rocks and become inoperable. NBV exhibits similarly poor performance in the maze. However, in this case, NBV's myopic planning is particularly ineffectual when faced with navigating a topologically-complex space, and the robot ultimately gets \textit{stuck}.   
As was the case in the subway, HFE suffers in the topologically-complex maze due to an accumulation of suboptimal local decisions. In particular, frontiers are sometimes not detected in the sharp bends of the maze, leaving the robot with an empty local policy space. As a result, the robot cannot progress and spends considerable time backtracking along the IRM to distant frontiers.

\subsection{Real-World Evaluation}
We extensively validated PLGRIM on physical robots in real-world environments. In particular, PLGRIM was run on Au-Spot in a lava tube, located in Lava Beds National Monument, Tulelake, CA. The cave consists of a main tube, which branches into smaller, auxiliary tubes. The floor is characterized by ropy masses of cooled lava. Large boulders, from ceiling breakdown, are scattered throughout the tube.
Fig.~\ref{fig:lava_tube_traj} shows the robot's trajectory overlaid on the aggregated LiDAR point cloud.
Fig.~\ref{fig:mlp_hardware_tests} and \ref{fig:glp_hardware_tests} discuss how PLGRIM is able to overcome the challenges posed by large-scale environments with complex terrain and efficiently guide the robot's exploration. Fig.~\ref{fig:all_together_plot}(f) shows the area covered over time.

\section{Conclusion}\label{sec:conclusion}
In this work, we develop a hierarchical framework for exploring large-scale, unknown environments with complex terrain in a POMDP setting. 
To obtain a tractable solution, we introduce a hierarchical belief space representation that effectively encodes a large-scale world state, while simultaneously capturing high-fidelity information local to the robot. Then we propose cascaded POMDP solvers that reason over long horizons within a suitable replanning time. 
We demonstrate our framework in high-fidelity dynamic simulation environments and in real-world environments, namely a natural cave.  
Future work includes incorporating semantic information gain, such as a science target signature, into the IRMs,
as well as extending the PLGRIM framework to multi-robot coverage problems.

\section*{Acknowledgments}
The work is partially supported by the Jet Propulsion Laboratory, California Institute of Technology, under a contract with the National Aeronautics and Space Administration (80NM0018D0004), Defense Advanced Research Projects Agency (DARPA), and the BRAILLE Mars Analog project funded by the NASA PSTAR program (NNH16ZDA001N).
We acknowledge our team members in Team CoSTAR for the DARPA Subterranean Challenge, Dr. Jennifer Blank at NASA Ames, Shushman Choudhury, and the resource staff at Lava Beds National Monument for their support.

{\small
\bibliographystyle{IEEEtran}  %
\bibliography{references_full}  %

\begin{thebibliography}{10}
\providecommand{\url}[1]{#1}
\csname url@samestyle\endcsname
\providecommand{\newblock}{\relax}
\providecommand{\bibinfo}[2]{#2}
\providecommand{\BIBentrySTDinterwordspacing}{\spaceskip=0pt\relax}
\providecommand{\BIBentryALTinterwordstretchfactor}{4}
\providecommand{\BIBentryALTinterwordspacing}{\spaceskip=\fontdimen2\font plus
\BIBentryALTinterwordstretchfactor\fontdimen3\font minus
  \fontdimen4\font\relax}
\providecommand{\BIBforeignlanguage}[2]{{%
\expandafter\ifx\csname l@#1\endcsname\relax
\typeout{** WARNING: IEEEtran.bst: No hyphenation pattern has been}%
\typeout{** loaded for the language `#1'. Using the pattern for}%
\typeout{** the default language instead.}%
\else
\language=\csname l@#1\endcsname
\fi
#2}}
\providecommand{\BIBdecl}{\relax}
\BIBdecl

\bibitem{blank2020robotic}
J.~G. Blank, ``Robotic mapping and exploration of a terrestrial lava tube: A
  structured planetary cave mission simulation with a remote astrobiology
  science team,'' in \emph{3rd International Planetary Caves Conference}, 2020.

\bibitem{nagatani2013emergency}
K.~Nagatani, S.~Kiribayashi, Y.~Okada, K.~Otake, K.~Yoshida, S.~Tadokoro,
  T.~Nishimura, T.~Yoshida, E.~Koyanagi, M.~Fukushima \emph{et~al.},
  ``Emergency response to the nuclear accident at the fukushima daiichi nuclear
  power plants using mobile rescue robots,'' \emph{Journal of Field Robotics},
  vol.~30, no.~1, pp. 44--63, 2013.

\bibitem{pomdps_monahan1982}
G.~E. Monahan, ``State of the art—a survey of partially observable markov
  decision processes: theory, models, and algorithms,'' \emph{Management
  Science}, vol.~28, no.~1, pp. 1--16, 1982.

\bibitem{KLC98}
L.~Kaelbling, M.~Littman, and A.~Cassandra, ``Planning and acting in partially
  observable stochastic domains,'' \emph{Artificial Intelligence}, vol. 101,
  pp. 99--134, 1998.

\bibitem{Pineau03}
J.~Pineau, G.~Gordon, and S.~Thrun, ``Point-based value iteration: An anytime
  algorithm for {POMDP}s,'' in \emph{International Joint Conference on
  Artificial Intelligence}, 2003, pp. 1025--1032.

\bibitem{silver2010monte}
D.~Silver and J.~Veness, ``Monte-{C}arlo planning in large {POMDP}s,'' in
  \emph{Advances in Neural Information Processing Systems}, 2010, pp.
  2164--2172.

\bibitem{somani2013despot}
A.~Somani, N.~Ye, D.~Hsu, and W.~S. Lee, ``{DESPOT}: Online {POMDP} planning
  with regularization,'' in \emph{Advances in Neural Information Processing
  Systems}, 2013, pp. 1772--1780.

\bibitem{bonet1998learning}
B.~Bonet and H.~Geffner, ``Learning sorting and decision trees with {POMDP}s.''
  in \emph{International Conference on Machine Learning}, 1998, pp. 73--81.

\bibitem{kim2019pomhdp}
S.-K. Kim, O.~Salzman, and M.~Likhachev, ``{POMHDP}: Search-based belief space
  planning using multiple heuristics,'' in \emph{Proceedings of the
  International Conference on Automated Planning and Scheduling}, vol.~29,
  2019, pp. 734--744.

\bibitem{Ali14-IJRR}
A.~{Agha-mohammadi}, S.~Chakravorty, and N.~Amato, ``{FIRM}: Sampling-based
  feedback motion planning under motion uncertainty and imperfect
  measurements,'' \emph{International Journal of Robotics Research}, vol.~33,
  no.~2, pp. 268--304, 2014.

\bibitem{yamauchi1997frontier}
B.~Yamauchi, ``A frontier-based approach for autonomous exploration,'' in
  \emph{IEEE International Symposium on Computational Intelligence in Robotics
  and Automation}, 1997, pp. 146--151.

\bibitem{tao2007motion}
T.~Tao, Y.~Huang, F.~Sun, and T.~Wang, ``Motion planning for slam based on
  frontier exploration,'' in \emph{International Conference on Mechatronics and
  Automation}, 2007, pp. 2120--2125.

\bibitem{keidar2012robot}
M.~Keidar and G.~A. Kaminka, ``Robot exploration with fast frontier detection:
  theory and experiments,'' in \emph{International Conference on Autonomous
  Agents and Multiagent Systems}, 2012, pp. 113--120.

\bibitem{heng2015efficient}
L.~Heng, A.~Gotovos, A.~Krause, and M.~Pollefeys, ``Efficient visual
  exploration and coverage with a micro aerial vehicle in unknown
  environments,'' in \emph{IEEE International Conference on Robotics and
  Automation}, 2015, pp. 1071--1078.

\bibitem{gonzalez2002navigation}
H.~H. Gonz{\'a}lez-Banos and J.-C. Latombe, ``Navigation strategies for
  exploring indoor environments,'' \emph{International Journal of Robotics
  Research}, vol.~21, no. 10-11, pp. 829--848, 2002.

\bibitem{grabowski2003autonomous}
R.~Grabowski, P.~Khosla, and H.~Choset, ``Autonomous exploration via regions of
  interest,'' in \emph{IEEE/RSJ International Conference on Intelligent Robots
  and Systems}, vol.~2, 2003, pp. 1691--1696.

\bibitem{pathak_icm}
D.~Pathak, P.~Agrawal, A.~A. Efros, and T.~Darrell, ``Curiosity-driven
  exploration by self-supervised prediction,'' in \emph{arXiv preprint
  arXiv:1705.05363}, 2017.

\bibitem{burda2018study}
Y.~Burda, H.~Edwards, D.~Pathak, A.~Storkey, T.~Darrell, and A.~Efros,
  ``Large-scale study of curiosity-driven learning,'' in \emph{arXiv preprint
  arXiv:1808.04355}, 2018.

\bibitem{rnd}
Y.~Burda, H.~A. Edwards, A.~J. Storkey, and O.~Klimov, ``Exploration by random
  network distillation,'' in \emph{arXiv preprint arXiv:1810.12894}, 2018.

\bibitem{ECR2018}
N.~Savinov, A.~Raichuk, R.~Marinier, D.~Vincent, M.~Pollefeys, T.~Lillicrap,
  and S.~Gelly, ``Episodic curiosity through reachability,'' in \emph{arXiv
  preprint arXiv:1810.02274}, 2018.

\bibitem{kurniawati2011motion}
H.~Kurniawati, Y.~Du, D.~Hsu, and W.~S. Lee, ``Motion planning under
  uncertainty for robotic tasks with long time horizons,'' \emph{International
  Journal of Robotics Research}, vol.~30, no.~3, pp. 308--323, 2011.

\bibitem{bai2015intention}
H.~Bai, S.~Cai, N.~Ye, D.~Hsu, and W.~S. Lee, ``Intention-aware online {POMDP}
  planning for autonomous driving in a crowd,'' in \emph{IEEE International
  Conference on Robotics and Automation}, 2015, pp. 454--460.

\bibitem{indelman2015planning}
V.~Indelman, L.~Carlone, and F.~Dellaert, ``Planning in the continuous domain:
  A generalized belief space approach for autonomous navigation in unknown
  environments,'' \emph{International Journal of Robotics Research}, vol.~34,
  no.~7, pp. 849--882, 2015.

\bibitem{martinez2009bayesian}
R.~Martinez-Cantin, N.~De~Freitas, E.~Brochu, J.~Castellanos, and A.~Doucet,
  ``A bayesian exploration-exploitation approach for optimal online sensing and
  planning with a visually guided mobile robot,'' \emph{Autonomous Robots},
  vol.~27, no.~2, pp. 93--103, 2009.

\bibitem{Lauri2016planning}
M.~Lauri and R.~Ritala, ``Planning for robotic exploration based on forward
  simulation,'' \emph{Robotics and Autonomous Systems}, vol.~83, pp. 15--31,
  2016.

\bibitem{kaelbling2011planning}
L.~P. Kaelbling and T.~Lozano-P{\'e}rez, ``Planning in the know: Hierarchical
  belief-space task and motion planning,'' in \emph{Workshop on Mobile
  Manipulation, IEEE International Conference on Robotics and Automation},
  2011.

\bibitem{umari2017autonomous}
H.~Umari and S.~Mukhopadhyay, ``Autonomous robotic exploration based on
  multiple rapidly-exploring randomized trees,'' in \emph{IEEE/RSJ
  International Conference on Intelligent Robots and Systems}, 2017, pp.
  1396--1402.

\bibitem{dang2019explore}
T.~Dang, S.~Khattak, F.~Mascarich, and K.~Alexis, ``Explore locally, plan
  globally: A path planning framework for autonomous robotic exploration in
  subterranean environments,'' in \emph{International Conference on Advanced
  Robotics}, 2019, pp. 9--16.

\bibitem{kim2019bi}
S.-K. Kim, R.~Thakker, and A.-A. Agha-Mohammadi, ``Bi-directional value
  learning for risk-aware planning under uncertainty,'' \emph{IEEE Robotics and
  Automation Letters}, vol.~4, no.~3, pp. 2493--2500, 2019.

\bibitem{vien2015hierarchical}
N.~A. Vien and M.~Toussaint, ``Hierarchical monte-carlo planning,'' in
  \emph{AAAI Conference on Artificial Intelligence}, 2015.

\bibitem{bircher2016receding}
A.~Bircher, M.~Kamel, K.~Alexis, H.~Oleynikova, and R.~Siegwart, ``Receding
  horizon ``next-best-view'' planner for {3D} exploration,'' in \emph{IEEE
  International Conference on Robotics and Automation}, 2016, pp. 1462--1468.

\bibitem{fan2021step}
D.~D. Fan, K.~Otsu, Y.~Kubo, A.~Dixit, J.~Burdick, and A.-a. Agha-mohammadi,
  ``{STEP}: {S}tochastic traversability evaluation and planning for safe
  off-road navigation,'' in \emph{arXiv preprint arXiv:2103.02828}, 2021.

\bibitem{Ebadi2020}
K.~Ebadi, Y.~Chang, M.~Palieri, A.~Stephens, A.~Hatteland, E.~Heiden,
  A.~Thakur, B.~Morrell, L.~Carlone, and A.~Agha-mohammadi, ``{LAMP}:
  Large-scale autonomous mapping and positioning for exploration of
  perceptually-degraded subterranean environments,'' in \emph{IEEE
  International Conference on Robotics and Automation}, 2020.

\bibitem{littman1995learning}
M.~Littman, A.~Cassandra, and L.~Kaelbling, ``Learning policies for partially
  observable environments: Scaling up,'' in \emph{Machine Learning
  Proceedings}, 1995, pp. 362--370.

\bibitem{Otsu2020}
K.~Otsu, S.~Tepsuporn, R.~Thakker, T.~S. Vaquero, J.~A. Edlund, W.~Walsh,
  G.~Miles, T.~Heywood, M.~T. Wolf, and A.~Agha-mohammadi, ``{Supervised
  autonomy for communication-degraded subterranean exploration by a robot
  team},'' in \emph{IEEE Aerospace Conference}, 2020.

\bibitem{AliNeBula21arXiv}
A.~Agha-mohammadi and {et al.}, ``{NeBula}: Quest for robotic autonomy in
  challenging environments; {TEAM CoSTAR} at the {DARPA} subterranean
  challenge,'' in \emph{arXiv preprint arXiv:2103.11470}, 2021.

\bibitem{AutoSpot}
A.~{Bouman$*$}, M.~{Ginting$*$}, N.~{Alatur$*$}, M.~Palieri, D.~Fan, T.~Touma,
  T.~Pailevanian, S.~Kim, K.~Otsu, J.~Burdick, and A.~Agha-Mohammadi,
  ``{Autonomous Spot: Long-Range Autonomous Exploration of Extreme Environments
  with Legged Locomotion},'' in \emph{IEEE/RSJ International Conference on
  Intelligent Robots and Systems}, 2020.

\end{thebibliography}
}

\end{document}